\documentclass[lettersize,journal]{IEEEtran}
\usepackage{amsmath,amsfonts}
\usepackage{algorithmic}
\usepackage{array}
\usepackage[caption=false,font=normalsize,labelfont=sf,textfont=sf]{subfig}
\usepackage{textcomp}
\usepackage{stfloats}
\usepackage{url}
\usepackage{verbatim}
\usepackage{graphicx}

\usepackage{amssymb}
\usepackage{arydshln}

\usepackage{abbr}
\usepackage{times}
\usepackage{epsfig}
\usepackage{caption}
\usepackage{algorithm}
\usepackage{bbm}
\usepackage{bm}
\usepackage{color}
\usepackage{multirow}
\usepackage{booktabs}
\usepackage{diagbox}
\usepackage[table]{xcolor}
\usepackage[pagebackref,breaklinks,colorlinks,linkcolor=black,anchorcolor=black,citecolor=black]{hyperref}
\usepackage{soul}
\sethlcolor{gray}

\newcommand{\Tref}[1]{Table~\ref{#1}}

\newcommand{\fref}[1]{Fig.~\ref{#1}}
\newcommand{\sref}[1]{Sec.~\ref{#1}}

\def\BibTeX{{\rm B\kern-.05em{\sc i\kern-.025em b}\kern-.08em
    T\kern-.1667em\lower.7ex\hbox{E}\kern-.125emX}}
\usepackage{balance}

\begin{document}
\title{PointCAT: Contrastive Adversarial Training for Robust Point Cloud Recognition}
\author{Qidong~Huang,~Xiaoyi~Dong,~Dongdong~Chen,~Hang~Zhou,~Weiming~Zhang,~Kui~Zhang,\\~Gang~Hua,~\IEEEmembership{Fellow,~IEEE},~and~Nenghai~Yu
\IEEEcompsocitemizethanks{
  \IEEEcompsocthanksitem Qidong Huang, Xiaoyi Dong, Weiming Zhang, Kui Zhang and Nenghai Yu are with School of Cyber Science and Security, University of Science and Technology of China, Hefei, Anhui 230026, China. E-mail: \{hqd0037@mail., dlight@mail., zhangwm@, zk19@mail.,  ynh@\}ustc.edu.cn
  \IEEEcompsocthanksitem Dongdong Chen is with Microsoft Research, Redmond, Washington 98052, USA. E-mail: cddlyf@gmail.com
  \IEEEcompsocthanksitem Hang Zhou is with Simon Fraser University, Burnaby, BC V5A1S6, Canada. E-mail: zhouhang2991@gmail.com
  \IEEEcompsocthanksitem Gang Hua is with Wormpex AI Research LLC, WA 98004, US, E-mail: ganghua@gmail.com
  \IEEEcompsocthanksitem Weiming Zhang is the corresponding author.
  }
}

\markboth{Journal of \LaTeX\ Class Files,~Vol.~18, No.~9, September~2020}%
{How to Use the IEEEtran \LaTeX \ Templates}

\maketitle


\begin{figure*}[!h]
\centering
\begin{minipage}{0.32\linewidth}
    \centering
    \includegraphics[width=1\linewidth]{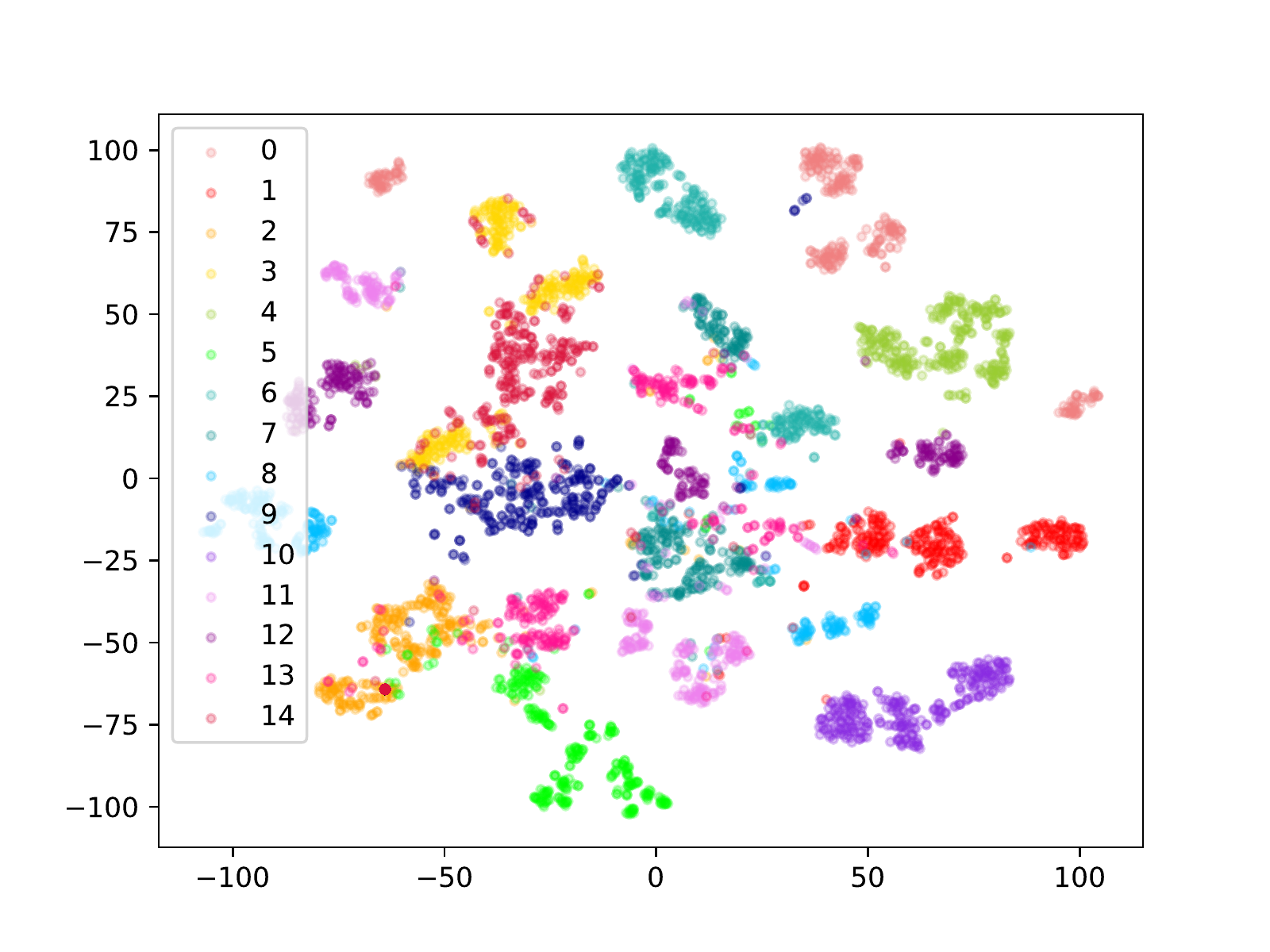}
    (a) t-SNE for vanilla AT
\end{minipage}
\hfill
\begin{minipage}{0.32\linewidth}
    \centering
    \includegraphics[width=1\linewidth]{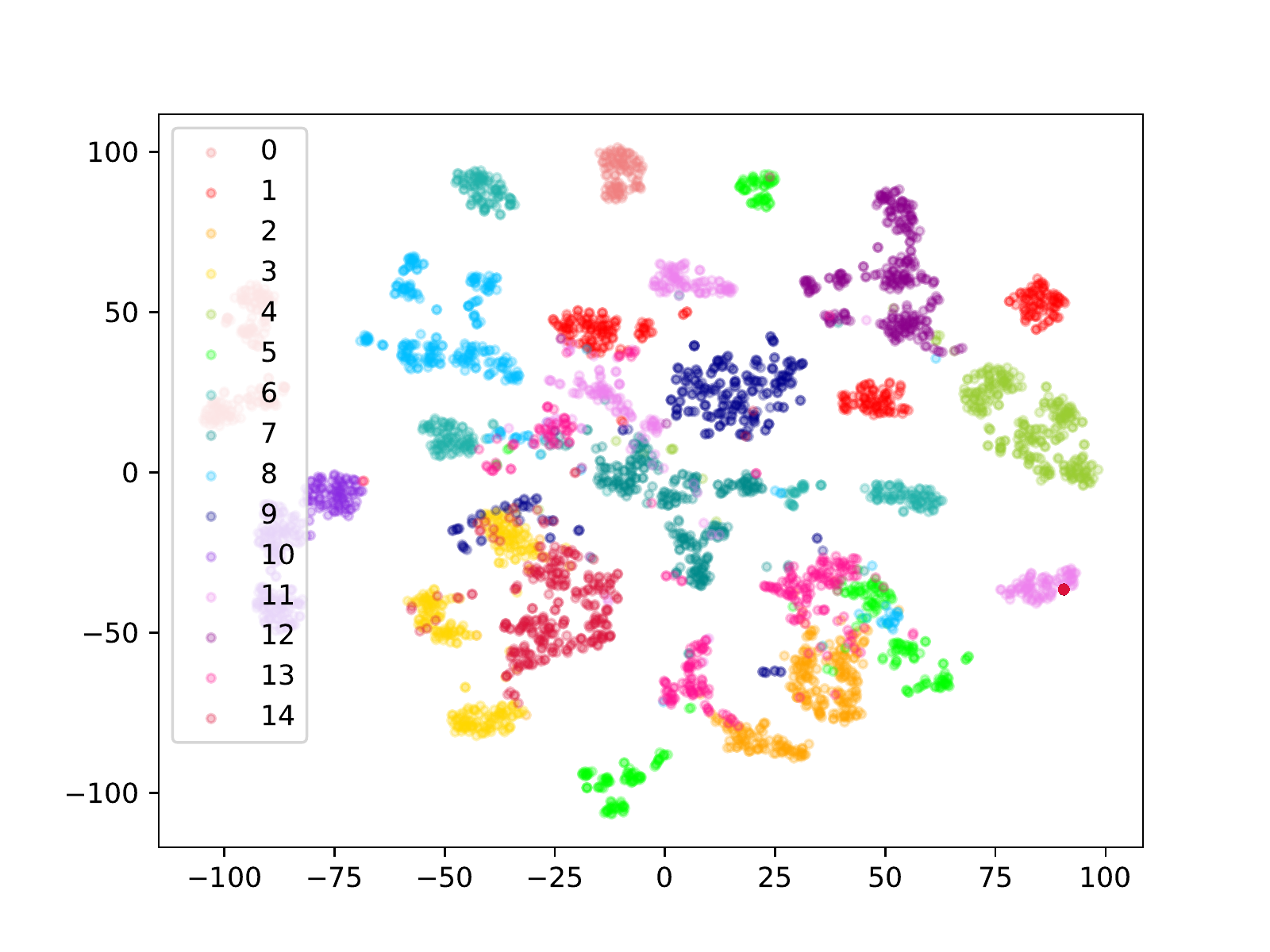}
    (b) t-SNE for TRADES
\end{minipage}
\hfill
\begin{minipage}{0.32\linewidth}
    \centering
    \includegraphics[width=1\linewidth]{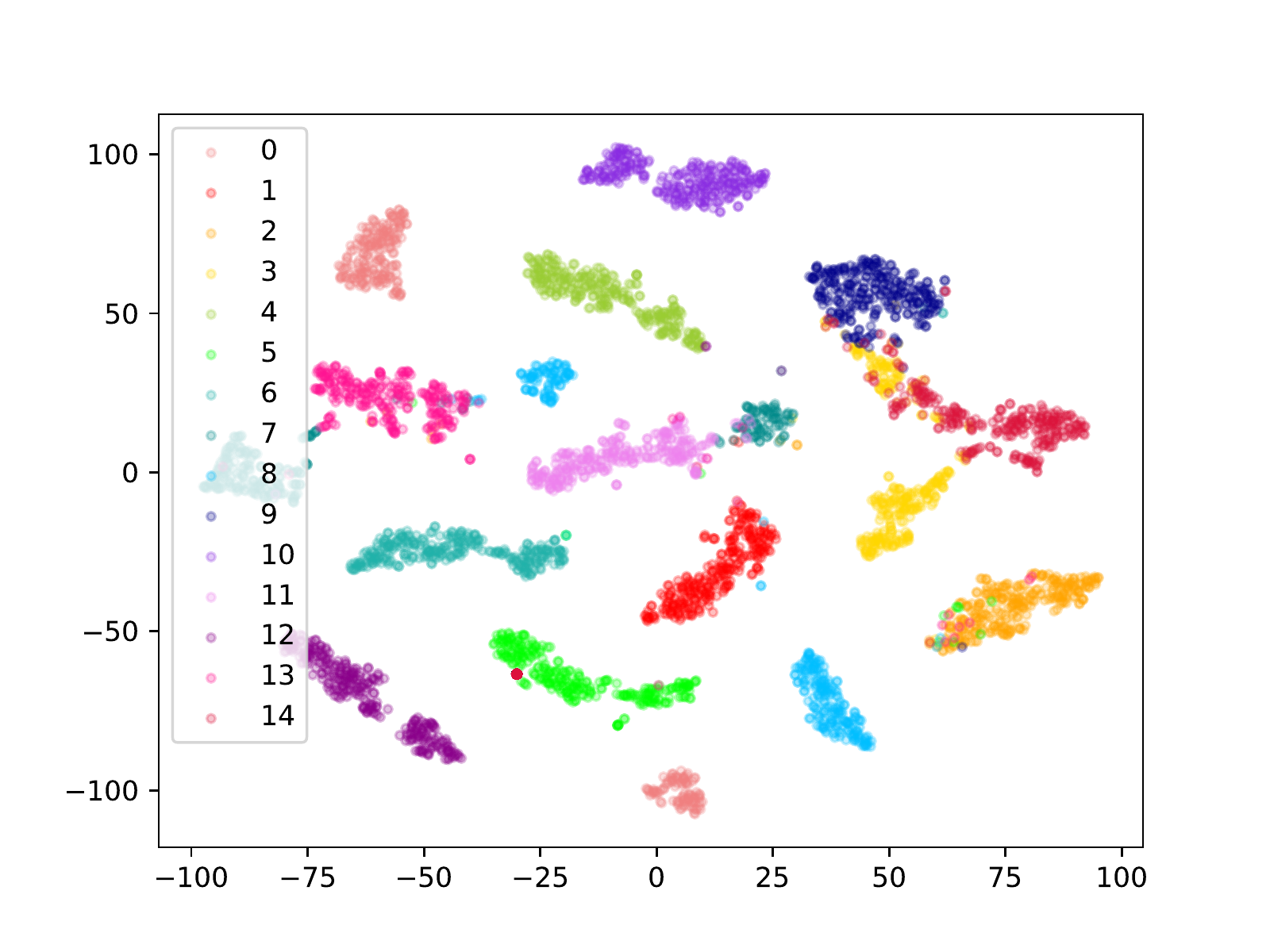}
    (c) t-SNE for PointCAT
\end{minipage}
\caption{Feature visualization on CurveNet \cite{xiang2021curvenet} trained by vanilla AT, TRADES and our PointCAT, respectively. The point cloud samples are randomly selected from 15 categories in ModelNet40 \cite{Wu2015modelnet} and subsequently corrupted as model inputs. Different numbers or colors corresponds to different categories. PointCAT learns the obviously more clustered and discriminative features.}
\label{fig:intro_1}
\end{figure*}

\begin{abstract}
Notwithstanding the prominent performance achieved in various applications, point cloud recognition models have often suffered from natural corruptions and adversarial perturbations. 
In this paper, we delve into boosting the general robustness of point cloud recognition models and propose Point-Cloud Contrastive Adversarial Training (PointCAT).
The main intuition of PointCAT is encouraging the target recognition model to narrow the decision gap between clean point clouds and corrupted point clouds. 
Specifically, we leverage a supervised contrastive loss to facilitate the alignment and uniformity of the hypersphere features extracted by the recognition model, and design a pair of centralizing losses with the dynamic prototype guidance to avoid these features deviating from their belonging category clusters.
To provide the more challenging corrupted point clouds, we adversarially train a noise generator along with the recognition model from the scratch, instead of using gradient-based attack as the inner loop like previous adversarial training methods.
Comprehensive experiments show that the proposed PointCAT outperforms the baseline methods and dramatically boosts the robustness of different point cloud recognition models, under a variety of corruptions including isotropic point noises, the LiDAR simulated noises, random point dropping and adversarial perturbations.
\end{abstract}


\section{Introduction}


\IEEEPARstart{P}{oint} clouds, captured by 3D sensors like LiDAR and Kinect, have become one of the most popular representations for depicting object surfaces and modeling 3D shapes. 
Its impressive performance has been witnessed in various applications (\eg, robotics, immersive tele-presence) and security-critical scenarios like autonomous driving, autopilot, \etc. 
With the purpose of precisely categorize 3D objects, point cloud recognition \cite{9483674,9668913} takes a basic and significant part in many downstream tasks such as point cloud analysis \cite{9292465,9405439} and object detection \cite{9234727,9792578,9429889}.

However, point cloud recognition is still confronted with many threats in practice. 
Due to the unpredictable environment and the inherent limitations of scanning equipment, it is inevitable for point cloud data to get perturbed and mutilated by real-world corruptions or reconstruction distortions from images. 
More crucially, recent works \cite{geoa3,Xiang3dadv,SunCCM20,TuRMLYDCU20} have demonstrated that point cloud recognition models are also susceptible to adversarial attacks especially adaptive attacks \cite{TramerCBM20}, which is attributed by the vulnerability of deep neural networks to imperceptible perturbations \cite{SzegedyZSBEGF13,carlini2017towards,dong18mifgm,XieZZBWRY19,HeZRS16,goodfellow2015explaining}. 
These corruptions seriously affects the accuracy of point cloud recognition, bringing a lot of security concerns and disputes on issues like autonomous driving. 
Accordingly, the general robustness of point cloud recognition still remains an imperative topic of nowadays study.

Existing works on robust point cloud recognition can be roughly divided into two categories. 
The first one is to preprocess the inputs and purify them into clean data, \eg, statistical outlier removal \cite{zhou2019dupnet}. 
But the involvement of such preprocess modules often leads to extra time cost, which dramatically degrades the recognition efficiency.
The other one is to enhance the recognition model itself with gather-vectors \cite{Dong2020gvg} or adversarial training methods \cite{Madry18adversarial,zhang2019trades,cui2020lbgat} that succeed in image tasks. 
Unfortunately, the robustness improvement provided by these methods is still limited for defending against both natural corruptions and adversarial attacks.

Through investigating into the commonality of previous adversarial training methods, we find that most of them focus on designing a particular loss function to construct the robust decision boundary \cite{zhang2019trades,cui2020lbgat}, but ignore the importance of learning the robust feature extraction. 
As t-SNE \cite{vandermaaten08a} visualized in \fref{fig:intro_1}, when we project both clean samples and corrupted samples onto the hypersphere space, the features learned by vanilla adversarial training (AT) \cite{Madry18adversarial} or TRADES \cite{zhang2019trades} are obviously not clustered enough, leading to more difficult decision boundary construction. 
By contrast, our method can learn the more category-wisely clustered and discriminative features, so that the subsequently classification will be much easier. 
From this perspective, we intuitively conclude the two objectives of robust feature extraction as: 
1) facilitating the category-wise alignment and uniformity on feature hypersphere; 
2) enabling the same categorized features to be concentrated.

Motivated by these, we propose a novel prototype-guided contrastive adversarial training method for robust point cloud recognition, named \textbf{``PointCAT''}. 
In the light of InfoNCE loss \cite{TschannenDRGL20} that performs well in contrastive learning \cite{chen2020simclr,Chen21mocov3,Grill20byol,Chen21simsiam}, we leverage its supervised variant \cite{Khosla20supcon} to mitigate the high-level discrepancy among the same categorized clean/corrupted point clouds towards the first objective mentioned above. 
For the second objective, we design a pair of centralizing losses for clean/corrupted point clouds respectively with the guidance of a set of specially defined prototypes. 
These prototypes can be regarded as the feature-level centres of different category clusters, which are dynamically optimized with a data-independent strategy during training. 

Besides the robust learning paradigm, we also provide a new way to generate the required point cloud corruptions. 
Considering that it is impossible to involve all kinds of corruptions into our training, we utilize an autoencoder-like generator as a learnable attacker to explore more challenging corruptions. 
A learnable attacker usually behaves in a more flexible way \cite{XiongH20,9807638}, since it does not follow a pre-defined attack setting or a fixed configuration adopted by previous adversarial training methods (\eg, PGD \cite{Madry18adversarial} inner loop for vanilla AT). 
Such flexibility guarantees the distribution diversity of synthetic corruptions, which is beneficial for broadening the learning scope of recognition model. 
Both the noise generator and recognition model are alternately updated from the scratch. 
Thus the corruptions generated by the noise generator is progressively difficult, which acts as a better teacher for recognition model and adapt it to the perturbations of various intensities.

We evaluate the efficiency of the proposed PointCAT on four point cloud object datasets for various recognition models, including PointNet \cite{charles2017pointnet}, PointNet++ \cite{charles2017pointnet++}, DGCNN \cite{wang2019dgcnn} and CurveNet \cite{xiang2021curvenet}. 
Experimental results show that our method can not only outperform previous point cloud defenses and advanced adversarial training methods that succeed in image tasks, but also dramatically boost the robustness against white-box attacks, black-box attacks, Auto-Attack (AA) \cite{croce2020autoattack} and simulated LiDAR noises. 
Furthermore, we present LiMN20, a new dataset for validating the point cloud recognition robustness under the LiDAR scanning scenario, which consists of 1,000 complicated point clouds sampled by Blensor simulation \cite{Gschwandtner11blensor} from different positions and angles. 

To summarize, our contributions are four-fold as below.

\begin{itemize}
    \item We propose PointCAT, a novel contrastive adversarial learning framework for robust point cloud recognition.

    \item To the best of our knowledge, we are the first to consider the point cloud model robustness against both natural corruptions and adversarial attacks.

    \item To implement this, we propose the dynamic prototype guidance for robust feature extraction and a learnable noise generator to derive the challenging corruptions. 

    \item Extensive evaluations on four datasets prove the superior performance of PointCAT. Besides, we contribute LiMN20, a new dataset regarding LiDAR-simulated point clouds that are scanned by HDL-64E2. 
    
\end{itemize}

\begin{figure*}[!h]
\centering
\includegraphics[width=1\linewidth]{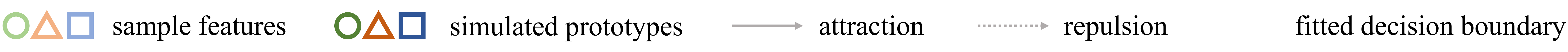}
\vfill
\begin{minipage}{0.31\linewidth}
    \centering
    \includegraphics[width=1\linewidth]{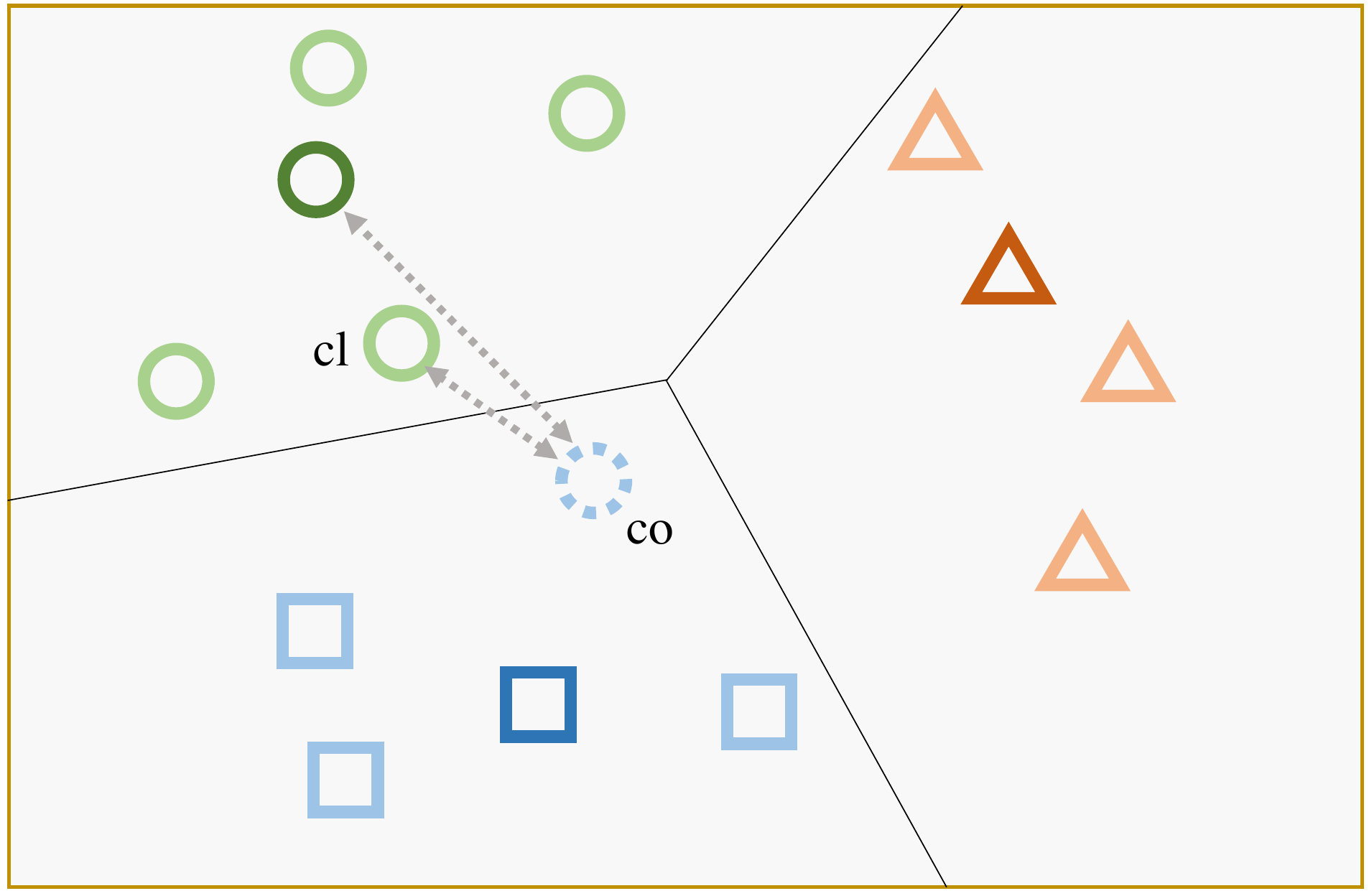}
    \\ 
    (a) Corrupted Sample Generation
\end{minipage}
\hfill
\begin{minipage}{0.31\linewidth}
    \centering
    \includegraphics[width=1\linewidth]{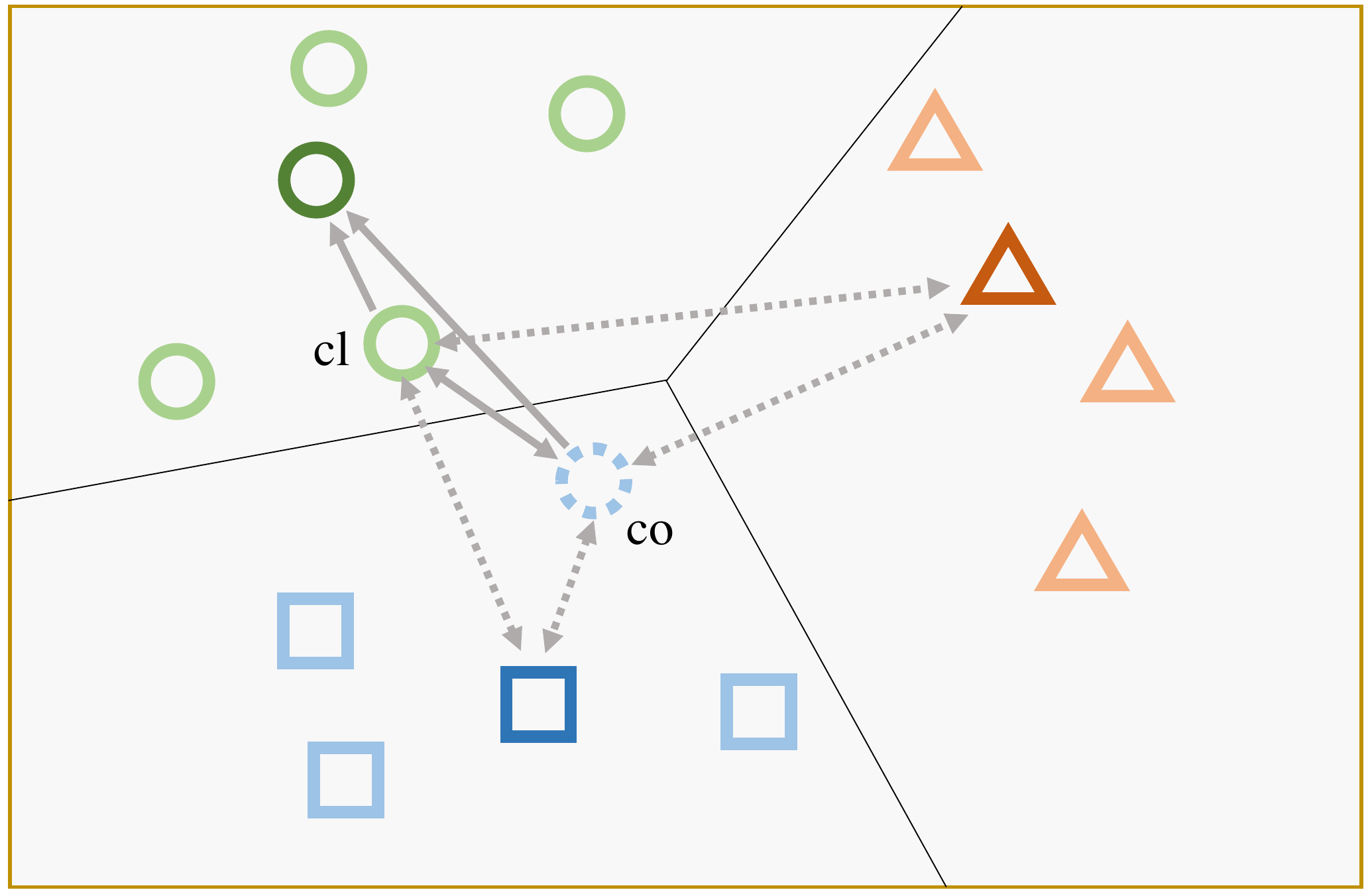}
    \\ 
    (b) Feature Contrast and Centralizing
\end{minipage}
\hfill
\begin{minipage}{0.31\linewidth}
    \centering
    \includegraphics[width=1\linewidth]{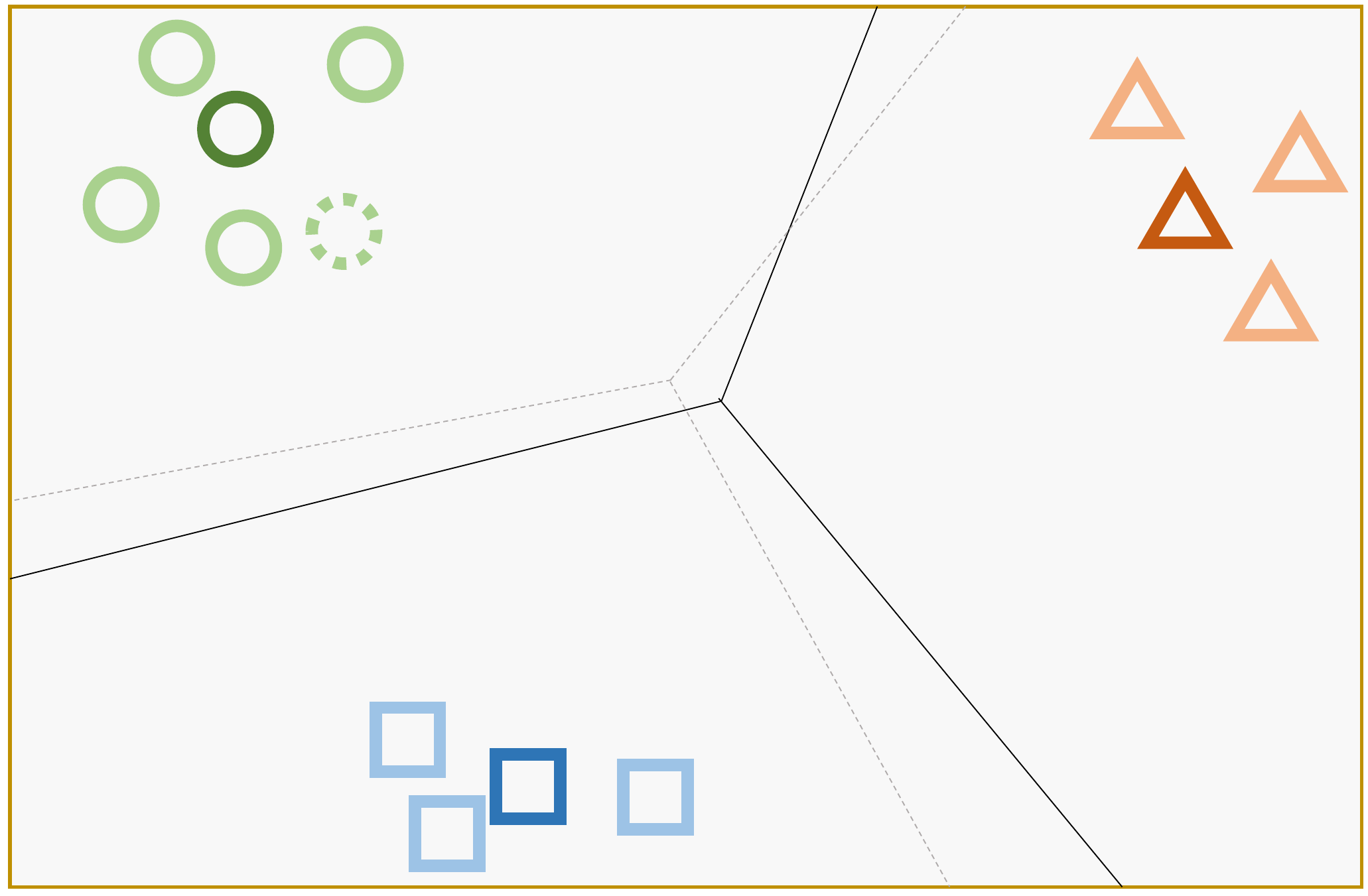}
    \\
    (c) After PointCAT training
\end{minipage}
\caption{Schematic diagram of PointCAT. Different colors denotes different categories. ``cl'' means the ``clean'' sample, and ``co'' means the ``corrupted'' sample. Taking the ``cl'' sample as an example, we first learn the noise generator by pulling the ``co'' sample away from ``cl'' sample and its prototype. Then we train the recognition model by narrowing the feature gap between ``cl'' and ``co'' samples, meanwhile centralizing them towards their prototype and pulling them away from other prototypes. Finally, the learned features are more clustered and the simulated prototypes are more evenly distributed on the hypersphere.}
\label{fig:intro_2}
\end{figure*}

\section{Related Work}

\subsection{Point Cloud Recognition}

Point cloud is one of the data formats to describe the scanned object surface, which is irregularly formed by a set of unordered and discrete points with 3D coordinates. 
Aiming to precisely recognize point cloud objects, various fundamental backbones or feature extracting strategies \cite{charles2017pointnet,charles2017pointnet++,Goyal21simpleview,Xu21paconv,ma2022rethinking,Ran_2022_CVPR,9709703} have been proposed in recent years. 
PointNet \cite{charles2017pointnet} is one of the pioneering works to directly utilize multi-layer perceptron (MLP) to extract point features and aggregate them by Maxpooling. 
And its variant, named PointNet++ \cite{charles2017pointnet++}, leverages the hierarchical structure to further improve the local feature extraction. 
With the well-designed neighborhood graphs, DGCNN \cite{wang2019dgcnn} is one of the most representative works which adopt convolutional networks to exploit point structures. 
Recently, CurveNet \cite{xiang2021curvenet} achieves the more satisfying classification accuracy on ModelNet40 through taking guided walks and aggregating hypothetical curves in point clouds. 
In this paper, we mainly implement PointCAT on the four aforementioned recognition models to demonstrate its efficacy.

\subsection{Adversarial Training}

Adversarial training, first proposed by Madry \etal \cite{Madry18adversarial}, is a well-known countermeasure that can effectively defend against input corruptions especially adversarial perturbations. 
There are a lot of works focusing on adversarial training, developing a variety of methods based on static ensemble models \cite{TramerKPGBM18}, hidden layer noise propagation \cite{9462815}, adversarial logit pairing and universal first-order adversary \cite{Madry18adversarial} (\ie, using PGD as the inner loop), \etc. 
Besides, Zhang \etal \cite{zhang2019trades} explored the trade-off between model adversarial robustness and natural accuracy by regularizing the model output from natural images and adversarial inputs. 
Cui \etal \cite{cui2020lbgat} further improved the PGD-based adversarial training by forcing the robust model to inherit the classification boundary of the clean model. 
Some of the most advanced adversarial training approaches, though originally designed for 2D image tasks, are implemented on 3D point clouds as the baselines in this paper.

More recently, some works \cite{9832540,JiangCCW20,KimTH20,GowalHOMK21} collaborate with contrast-related paradigms to strengthen the robustness of image networks. 
However, like original contrastive learning, most of these works still relays on strong image data augmentation. 
But image and point cloud are totally different data formats. 
Image is continuous signals with dense semantic, thus strong augmentation will not change its semantic.
On the contrary, point cloud (single object for recognition) is a sparse point set with limited semantic, strong augmentation would change its semantic so it is hard to design proper augmentation for point cloud constrastive learning. 
To tackle with this, this paper propose a learnable noise generator to online provide augmented corruptions, which generate challenging positive pairs and force the model to learn robust semantic features.

\subsection{3D Adversarial Attack and Defense}

\subsubsection{Point Cloud Adversarial Attack}
The topic of adversarial attack and defense on point cloud recognition has drawn increasing attentions from both industry and academia. 
Xiang \etal \cite{Xiang3dadv} introduced optimization-based attack C\&W \cite{carlini2017towards} to generate adversarial 3D point clouds and realize white-box attack. 
Zhou \etal \cite{Zhou2020lggan} proposed a flexible label-guided framework to optimize targeted adversarial point clouds. 
Compared to the 2D counterparts, 3D adversarial attack often shows less transferability due to the very difference among point cloud recognition models, thus Hamdi \etal \cite{HamdiRTG20advpc} introduced autoencoder-based reconstruction into optimization to improve the adversarial transferability. 
To guarantee the surface-level smoothness and fairness, the geometry-aware methods like $GeoA^3$ \cite{geoa3}, shape constrained methods like SiAdv \cite{Huang_2022_CVPR} and ITA \cite{9839597} are designed to boost the attack imperceptibility. 
To perform the more practical adversarial attack in autonomous driving, Cao \etal \cite{CaoXCZPRCFM19} and Sun \etal \cite{SunCCM20,SunCCYAMX21} systematically implemented point cloud attack to the LiDAR perception module and demonstrated the security threats brought by 3D adversarial objects. 
Considering current autonomous driving is often equipped with multi-sensor fusion perception, a new physical-world attack \cite{CaoWXYFYCLL21} has been proposed to successfully fool both the camera and LiDAR . 

\subsubsection{Point Cloud Adversarial Defense}
How to effectively defend against the aforementioned attacks is still a under-researched problem. 
More and more point cloud defense solutions have been presented to alleviate this situation. 
By eliminating statistical outliers and upsampling point clouds, Zhou \etal \cite{zhou2019dupnet} were the first to propose the pluggable preprocess module SOR and DUP-Net for point cloud defense. 
Dong \etal \cite{Dong2020gvg} proposed GvG through calculating gather-vectors to indicate global center and checking if it deviates from the normal region. 
But unfortunately, current countermeasures including adversarial training are still ineffective or time-consuming against both natural corruptions and adversarial perturbations. 
Hence it is significant to find a more general framework that can strengthen the overall robustness of 3D recognition models.

\begin{figure*}[!h]
\centering
\includegraphics[width=1.0\linewidth]{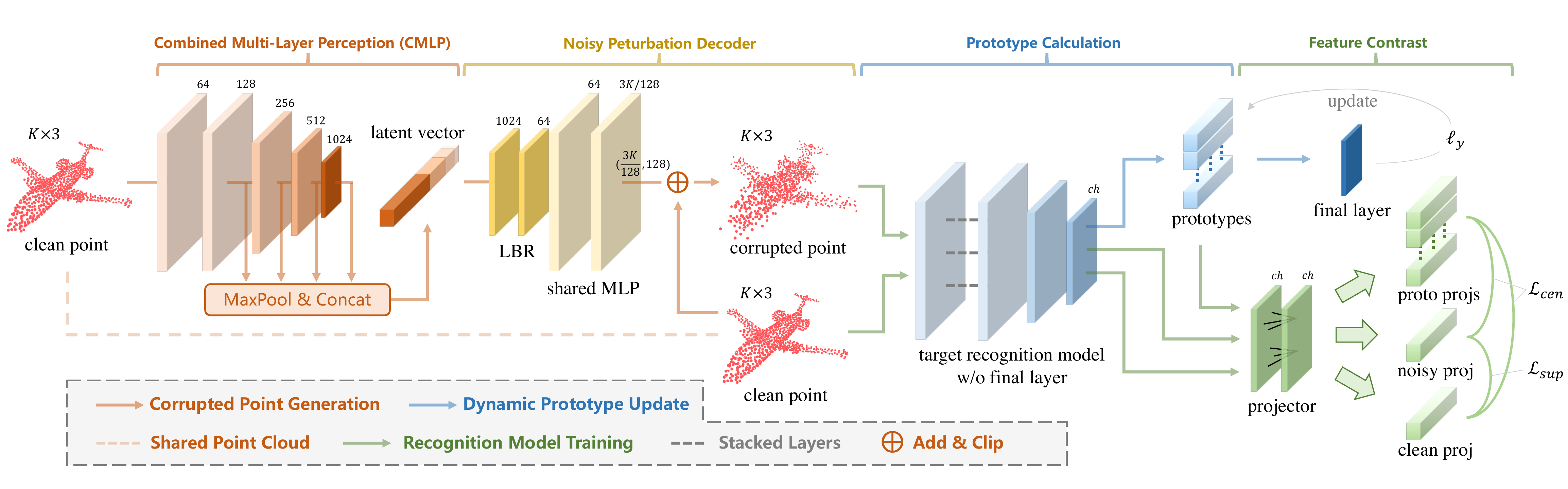}
\caption{The pipeline of PointCAT. \textbf{Step 1:} (blue arrow) we regularly update all of prototypes by maximizing $\ell_y$ before each batch training. \textbf{Step 2:} (orange arrow) we minimize loss $\mathcal{L}_{adv}$ (Eq.(\ref{eq:adv_loss})) to update noise generator $\mathcal{G}$ (consists of CMLP and Noisy Perturbation Decoder) for fixed iterations, then add the generated noise to clean input $x$ after the $l_2$-norm constraint to get corrupted point cloud $x'$. \textbf{Step 3:} (green arrow) we train recognition encoder with projector $\mathcal{P}\circ\mathcal{M}_e$ by minimizing $\mathcal{L}_{robust}$ (Eq.(\ref{eq:robust_loss})). The final layer $\mathcal{M}_f$ is trained alone in the end of each epoch training. LBR means Linear, BatchNorm and Relu.}
\label{fig:pipeline_pointcat}
\end{figure*}

\section{Proposed Method}

We first provide an overview of PointCAT framework. 
Then, we introduce the implementation details about the proposed method, including network design, objective loss function, important mechanisms and the algorithm pseudocode. 
Overall, the simplified training procedure is illustrated in \fref{fig:intro_2}.

\subsection{Overview}

The main idea of PointCAT is encouraging the recognition model to narrow the decision gap between clean example $x$ and adversarial corrupted example $x'$, so that the recognition model can obtain the robustness on such corruption. 
During the training phase of recognition model, we adversarially search the perturbation with an autoencoder-like generator $\mathcal{G}$ to acquire the more challenging $x'$.
To avoid both clean and corrupted examples deviating from their ground-truth category clusters in high-level dimensional space, the prototypes of all classification categories are optimized to limit both clean and noisy features within their belonging clusters. 
The overall pipeline of PointCAT has been illustrated in \fref{fig:pipeline_pointcat}.

Similar with the standard adversarial training proposed by Madry \etal \cite{Madry18adversarial}, our method also follows the adversarial game principle. Both noise generator $\mathcal{G}$ and recognition model $\mathcal{M}$ are trained from the scratch, where the former acts as the adversary and the latter performs as the defender. 
Specifically, $\mathcal{G}$ intends to explore the feature-level weakness of $\mathcal{M}$ and produces the more difficult corrupted example by $x' = x + \sigma (\mathcal{G} (x))$, where $\sigma$ denotes the $l_2$-norm ball to constrain the perturbation.
Conversely, $\mathcal{M}$ delves itself into contrasting the projection of corrupted example with the clean one in a supervised way. 
The training objective can be formulated as the following optimization problem:
\begin{equation}
\begin{split}
    \min_\theta \mathcal{L}_{robust} (\mathcal{P}\circ\mathcal{M}_e (x), \mathcal{P}\circ\mathcal{M}_e (x'), y), 
    \\
    \text{s.t.}\quad \min_\phi \mathcal{L}_{adv} (\mathcal{P}\circ\mathcal{M}_e (x), \mathcal{P}\circ\mathcal{M}_e (x'), y),
\end{split}
\end{equation}
where $\circ$ means the composite mapping applied to serially connect the recognition encoder $\mathcal{M}_e$ with a projection head $\mathcal{P}$, as suggested in the famous contrastive learning method SimCLR \cite{chen2020simclr}.
Recognition encoder with projector $\mathcal{P}\circ\mathcal{M}_e$ and noise generator $\mathcal{G}$ are parameterized by $\theta$ and $\phi$, respectively. 
Note that ground-truth label $y$ is used to get the prototype of its corresponding category, and the projector will be abandoned when the whole training gets completed. 

\subsection{Dynamic Prototype Optimizing}

At the beginning of each batch training phase, we need to update the estimated prototypes for all of $M$ classification categories (\eg, airplane, bench). 
Different from the solution used in deep clustering \cite{caron20swav,caron18deepclustering}, we adopt a data-independent strategy to search these prototypes since adversarial training is usually conducted in the supervised way. 
From this perspective, we investigate the original recognition model $\mathcal{M}$ (\ie, $\mathcal{M}_f\circ\mathcal{M}_e$, in which $\mathcal{M}_f$ refers to the final layer of the model) through calculating the following logit classification loss \cite{carlini2017towards} without max-margin:
\begin{equation}
    \ell_y (\psi) = \left[\mathcal{M} (\psi)\right]_y - \max_{t \neq y} \left[\mathcal{M} (\psi)\right]_t.
\end{equation}

Here $y$ is one of any category labels and $\left[\mathcal{M} (\psi)\right]_t$ represents the predicted probability score of label $t$, while $\psi$ is the input variable that we intend to optimize. 
Our intuition is that, the larger the score difference between $y$ and the second highest category is, the closer the estimated prototype is to the intra-class centre. 
By maximizing the logit loss $\ell_y$ for each category $y$, we can get the expected prototype $c_y$ with the recognition encoder, \ie, 
\begin{equation}
    c_y \triangleq \mathcal{M}_e (\psi_y^\ast), \quad \text{s.t.}\quad \psi_y^\ast = \mathop{\arg\max}_\psi \ell_y (\psi).
\label{eq:3}
\end{equation}

Before the whole adversarial training procedure, all of $\psi$ are initialized as random Gaussian noise points. 
For each batch training phase of PointCAT, we spend a few but fixed steps to maximize this logit loss and update prototypes iteratively. 
In this way, the evolution of prototypes and the recognition encoder can be always synchronized and matched.

\subsection{Adversarial Noise Generation}

Given a batch of clean point clouds $\left\{x_k\right\}_{k=1}^N$ where $x \in \mathbb{R}^{K\times3}$, \ie, each one consists of total $K$ points, we first leverage Combined Multi-Layer Perception (CMLP) \cite{Huang20pfnet} to encode them into a set of latent vectors. 
Compared with original PointNet encoder \cite{charles2017pointnet}, the channel-wise hierarchical design enables CMLP can absorb the information from both shallow-level and high-level features, which alleviates the strong effect of symmetric function Maxpooling. 
Then the latent vectors are decoded by two fully connected layers and two Multi-Layer Perception (MLP) modules. 
After reshaping them into the input shape, we constrain the perturbations into $l_2$-norm ball, and add them on original point clouds to obtain adversarial corrupted point cloud by
\begin{equation}
    x_k' = x_k + \sigma (\mathcal{G} (x_k)), \quad k = 1,\cdots,N,
\end{equation}
where $\mathcal{G}$ denotes the whole autoencoder-like noise generator.
In order to generate the more challenging $x_k'$, the training objective of $\mathcal{G}$ is 
1) to enlarge the \textbf{gap} between $x_k'$ and $x_k$, 
2) to find $x_k'$ which \textbf{escapes} as far away from its prototype $c_{y_k}$ as possible. 
Here $y_k$ is the ground-truth label of both $x_k$ and $x_k'$. 
Let $\left\{z_i\right\}_{i=1}^{2N}$ be all of total $2N$ projections calculated from clean and noisy features, \ie, $\forall k = 1,\cdots,N$,
\begin{equation}
\begin{split}
    z_{2k-1} = \mathcal{P}\circ\mathcal{M}_e (x_k), 
    \\
    z_{2k} = \mathcal{P}\circ\mathcal{M}_e (x_k'), 
\end{split}
\end{equation}
the objective function $\mathcal{L}_{adv}$ can be formulated as a weighted combination of the following two parts:
\begin{equation}
\begin{split}
    \mathcal{L}_{gap} = \frac{1}{N} \sum_{k=1}^N - \exp \left(-\text{sim} (z_{2k-1}, z_{2k}) / \tau_{adv} \right),
    \\
    \mathcal{L}_{esc} = \frac{1}{N} \sum_{k=1}^N -\exp \left(-\text{sim} (z_{2k}, \omega_{y_k}) / \tau_{adv} \right),
\end{split}
\end{equation}
in which $\text{sim} (\cdot, \cdot)$ denotes the cosine similarity function, symmetrically, $\text{sim} (u, v) = u \cdot v / |u| |v|$, and $\tau_{adv}$ is a temperature hyper-parameter.
$\omega_{y_k}$ is the projection of prototype, \ie, $\omega_{y_k} = \mathcal{P} (c_{y_k})$.
Therefore, the whole adversarial loss function for generator $\mathcal{G}$ training is 
\begin{equation}
    \mathcal{L}_{adv} = \mathcal{L}_{gap} + \beta \mathcal{L}_{esc}, 
\label{eq:adv_loss}
\end{equation}
where $\beta$ controls the weight of escaping loss component.

\subsection{Feature Contrast and Centralizing}

When the corrupted point generation in each batch training is completed, we consider to update the parameter of the recognition encoder with projector $\mathcal{P}\circ\mathcal{M}_e$. 
Specifically, we regard all batch samples that belong to category $y_k$ except $x_k$ is the positive set of $x_k$, while the remaining samples form the negative set. 
Note that both clean projection $z_{2k-1}$ and noisy projection $z_{2k}$ belong to the same category $y_k$.
Thus the indices of the positive set corresponding to projection $z_i$ can be defined as $P(i) \equiv \left\{j \big| j \in \{1,\cdots,2N \} / \{ i \},  y_{\lceil j/2 \rceil} = y_{\lceil i/2 \rceil} \right\}$.
Apparently, $P(2k-1)/\{2k\} = P(2k)/\{2k-1\}$.
To facilitate the alignment and uniformity of hypersphere features, we adopt supervised contrastive loss \cite{Khosla20supcon} to narrow the decision gap among positive set examples, \ie, $\forall i = 1,\cdots,2N$,
\begin{equation}
    \ell_i^{sup} = \sum_{j=1}^{2N} \frac{- \mathbbm{1}_{j \in P(i)}}{\left| P(i) \right|} \log \frac{\exp \left(\text{sim} (z_i, z_j) / \tau_{sup} \right)}{\sum_{s=1}^{2N} \mathbbm{1}_{s \neq i} \exp \left(\text{sim} (z_i, z_s) / \tau_{sup} \right)},
\end{equation}
where $\mathbbm{1}$ refers to the characteristic function and $\tau_{sup}$ is a temperature hyper-parameter.
$\left| P(i) \right|$ means the total number of examples in this positive set.

However, it is not enough to make model $\mathcal{M}$ become a robust learner if only using contrastive loss during the training phase. To avoid both the learned representations of $x_k$ and $x_k'$ deviating from their ground-truth category cluster, we draw them towards their prototype $c_{y_k}$ and pull them away from other prototypes meanwhile, by introducing a pair of centralizing loss for $x_k$ and $x_k'$ respectively:
\begin{equation}
\begin{split}
    \ell_k^{ori} = - \log \frac{\exp \left(\text{sim} (z_{2k-1}, \omega_{y_k}) / \tau_{cen} \right)}{\sum_{t=1}^M \mathbbm{1}_{t \neq y_k} \exp \left(\text{sim} (z_{2k-1}, \omega_t) / \tau_{cen} \right)},
    \\
    \ell_k^{adv} = - \log \frac{\exp \left(\text{sim} (z_{2k}, \omega_{y_k}) / \tau_{cen} \right)}{\sum_{t=1}^M \mathbbm{1}_{t \neq y_k} \exp \left(\text{sim} (z_{2k}, \omega_t) / \tau_{cen} \right)},
\end{split}
\end{equation}
where $\tau_{cen}$ is another temperature hyper-parameter. With the guidance of prototypes, the recognition model $\mathcal{M}$ can avoid model collapse and maintain the high recognition accuracy. Finally, we calculate the average value of supervised contrastive loss and centralizing loss across the whole batch. 
Overall, the robust loss function of $\mathcal{M}$ is defined as:
\begin{equation}
    \mathcal{L}_{sup} = \frac{1}{2N} \sum_{i=1}^{2N} \ell_i^{sup}, \mathcal{L}_{cen} = \frac{1}{N} \sum_{k=1}^N \left(\ell_k^{ori} + \ell_k^{adv} \right),
\end{equation}
\begin{equation}
    \mathcal{L}_{robust} = \mathcal{L}_{sup} + \alpha \mathcal{L}_{cen}.
\label{eq:robust_loss}
\end{equation}
where $\alpha$ balances the weight of centralizing loss.

\subsection{The Pseudocode of the Proposed Method}

We provide the detailed algorithm pseudocode in Algo \ref{alg:Algo1}. 
Before each batch training, we spend $T_1$ iterations to regularly update all prototypes where $T_1$ is a configurable number related to learning rate. 
The each batch training consists of adversarial noise generation (the inner loop) and recognition model training (the outer loop). 
After each epoch training, the model final layer $\mathcal{M}_f$ is trained alone with cross-entropy loss. 
When the training procedure converges, we can derive a recognition model robust to various input corruptions.

\begin{algorithm}[h!]
    \caption{Point-Cloud Contrastive Adversarial Training}
    \label{alg:Algo1}
        \begin{algorithmic}
            \STATE \textbf{Input: } recognition model $\mathcal{M} = \mathcal{M}_f\circ\mathcal{M}_e$, noise generator $\mathcal{G}$ parameterized by $\phi$, model encoder with projector $\mathcal{P}\circ\mathcal{M}_e$ parameterized by $\theta$, $\mathcal{M}_f$ parameterized by $\theta_f$, prototype update iterations $T_1$, inner loop number $T_2$, hyper-parameter $\alpha$, $\beta$, $\eta_1$ and $\eta_2$.
            \STATE \textbf{Output: } robust recognition model $\mathcal{M}_{r}$.
            \STATE Initialize $\mathcal{M}$, $\mathcal{G}$, $\mathcal{P}$ randomly; 
            \STATE Initialize prototype inputs $\psi_y$ with warm start; 
            \REPEAT 
            \STATE Sample batch data $\{x_k\}_{k=1}^N$ with labels $\{y_k\}_{k=1}^N$;
            \FOR{$t=1$ \textbf{to} $T_1$}
            \STATE Update $\psi_y = \psi_y + \eta_1\nabla_\psi \ell_y(\psi_y)$; \hfill \textit{\# Eq.(\ref{eq:3})}
            \ENDFOR
            \STATE Compute $c_y = $ $\mathcal{M}_e$ $(\psi_y)$; \hfill \textit{\# update prototypes}
            \FOR{$t=1$ \textbf{to} $T_2$}
            \STATE Compute $\mathcal{L}_{adv} = \mathcal{L}_{gap} + \beta \mathcal{L}_{esc}$; \hfill \textit{\# Eq.(\ref{eq:adv_loss})}
            \STATE Update $\phi = \phi - \eta_2\nabla_\phi \mathcal{L}_{adv}$; \hfill \textit{\# update }$\mathcal{G}$
            \ENDFOR
            \STATE Compute $\mathcal{L}_{robust} = \mathcal{L}_{sup} + \alpha \mathcal{L}_{cen}$; \hfill \textit{\# Eq.(\ref{eq:robust_loss})}
            \STATE Update $\theta = \theta - \eta_2\nabla_\theta\mathcal{L}_{robust}$; \hfill \textit{\# update }$\mathcal{P}\circ\mathcal{M}_e$
            \IF{\text{end of epoch}}
            \STATE Update $\theta_f = \theta_f - \eta_2\nabla_{\theta_f} CE (x, y)$; \hfill \textit{\# update }$\mathcal{M}_f$
            \ENDIF
            \UNTIL training converges
        \end{algorithmic}
\end{algorithm}

\section{Experiment}

\subsection{Experimental Setup}

\begin{table*}[t]
	\footnotesize
    \centering
    \caption{Quantitative comparison about regular white-box robustness on ModelNet40, tested on PointNet (a.k.a, PN), PointNet++ (a.k.a, PN++), DGCNN and CurveNet. ``None'' denotes the recognition model without any defense. ``Acc'' means the accuracy on clean point clouds, higher is better. ``ASR'' means the attack success rate, lower is better.}
    \label{tab:table1}
    \setlength{\tabcolsep}{1mm}{
    \begin{tabular}{c|l|c|p{11mm}<{\centering}p{11mm}<{\centering}p{11mm}<{\centering}p{11mm}<{\centering}p{11mm}<{\centering}|p{11mm}<{\centering}p{11mm}<{\centering}p{11mm}<{\centering}p{11mm}<{\centering}p{11mm}<{\centering}}
        \toprule
        \multicolumn{1}{c|}{\multirow{3}{*}{Model}}
        & \multicolumn{1}{c|}{\multirow{3}{*}{Defense}}
        & \multicolumn{1}{c|}{\multirow{3}{*}{Acc (\%)}} 
        & \multicolumn{5}{c|}{ASR (Targeted Attack) (\%)} 
        & \multicolumn{5}{c}{ASR (Untargeted Attack) (\%)}
        \\
        & &
        & FGM
        & IFGM 
        & MIFGM 
        & PGD 
        & C\&W 
        & FGM 
        & IFGM 
        & MIFGM 
        & PGD 
        & C\&W 
        \\
        & &
        & \cite{goodfellow2015explaining}
        & \cite{gu2015towards}
        & \cite{dong18mifgm}
        & \cite{Madry18adversarial}
        & \cite{carlini2017towards}
        & \cite{goodfellow2015explaining}
        & \cite{gu2015towards}
        & \cite{dong18mifgm}
        & \cite{Madry18adversarial}
        & \cite{carlini2017towards}
        \\
        \midrule
        \multirow{6}{*}{PN \cite{charles2017pointnet}}
        & \textcolor{gray!75}{None} & \textcolor{gray!75}{89.32} & \textcolor{gray!75}{4.78} & \textcolor{gray!75}{99.11} & \textcolor{gray!75}{98.22} & \textcolor{gray!75}{99.31} & \textcolor{gray!75}{99.96} & \textcolor{gray!75}{83.18} & \textcolor{gray!75}{99.84} & \textcolor{gray!75}{99.84} & \textcolor{gray!75}{99.96} & \textcolor{gray!75}{100.00} 
        \\
        & SOR \cite{zhou2019dupnet} & 87.58 & 4.21 & 73.46 & 48.99 & 85.37 & 99.19 & \textbf{34.40} & 69.41 & \textbf{65.60} & 71.27 & 99.23
        \\
        \cmidrule{2-13}
        & AT \cite{Madry18adversarial} & 87.44 & 3.53 & 79.34 & 79.29 & 82.66 & 94.73 & 54.09 & 78.16 & 84.40 & 82.62 & 93.40
        \\
        & TRADES \cite{zhang2019trades} & 86.59 & 3.81 & 86.79 & 85.49 & 89.14 & 98.74 & 58.91 & 87.44 & 92.18 & 90.48 & 98.70
        \\
        & LBGAT \cite{cui2020lbgat} & 81.69 & 3.36 & 49.88 & 47.61 & 53.20 & 67.14 & 65.24 & 84.36 & 86.47 & 86.22 & 88.57 
        \\
        & \textbf{Ours} & \colorbox{blue!20}{\textbf{87.97}} & \colorbox{blue!20}{\textbf{2.76}} & \colorbox{blue!20}{\textbf{21.35}} & \colorbox{blue!20}{\textbf{20.91}} & \colorbox{blue!20}{\textbf{24.07}} & \colorbox{blue!20}{\textbf{36.10}} & 37.60 & \colorbox{blue!20}{\textbf{65.76}} & 66.09 & \colorbox{blue!20}{\textbf{67.99}} & \colorbox{blue!20}{\textbf{85.94}} 
        \\
        \midrule
        \midrule
        \multirow{6}{*}{PN++ \cite{charles2017pointnet++}}
        & \textcolor{gray!75}{None} & \textcolor{gray!75}{91.94} & \textcolor{gray!75}{2.92} & \textcolor{gray!75}{91.17} & \textcolor{gray!75}{85.94} & \textcolor{gray!75}{92.02} & \textcolor{gray!75}{99.72} & \textcolor{gray!75}{56.40} & \textcolor{gray!75}{89.34} & \textcolor{gray!75}{89.63} & \textcolor{gray!75}{92.91} & \textcolor{gray!75}{100.00} 
        \\
        & SOR \cite{zhou2019dupnet} & \textbf{91.56} & 3.65 & 64.29 & 47.97 & 79.46 & 28.98 & 26.46 & 59.01 & 59.66 & 69.81 & \textbf{35.88}
        \\
        \cmidrule{2-13}
        & AT \cite{Madry18adversarial} & 86.91 & 2.88 & 49.55 & 38.94 & 50.36 & 31.60 & 26.62 & 69.37 & 54.94 & 74.31 & 60.45 
        \\
        & TRADES \cite{zhang2019trades} & 89.38 & 3.08 & 47.77 & 43.23 & \textbf{47.97} & 29.37 & 40.19 & 64.83 & 73.14 & 71.76 & 36.91 
        \\
        & LBGAT \cite{cui2020lbgat} & 78.19 & 2.92 & 40.83 & 35.89 & 53.83 & 31.91 & 40.64 & 68.80 & 66.77 & 72.00 & 51.22
        \\
        & \textbf{Ours} & 91.33 & \colorbox{blue!20}{\textbf{2.76}} & \colorbox{blue!20}{\textbf{34.93}} & \colorbox{blue!20}{\textbf{31.04}} & 49.51 & \colorbox{blue!20}{\textbf{28.24}} & \colorbox{blue!20}{\textbf{22.24}} & \colorbox{blue!20}{\textbf{49.35}} & \colorbox{blue!20}{\textbf{54.46}} & \colorbox{blue!20}{\textbf{53.53}} & 40.32 
        \\
        \midrule
        \midrule
        \multirow{6}{*}{DGCNN \cite{wang2019dgcnn}}
        & \textcolor{gray!75}{None} & \textcolor{gray!75}{92.34} & \textcolor{gray!75}{2.76} & \textcolor{gray!75}{97.04} & \textcolor{gray!75}{67.75} & \textcolor{gray!75}{92.26} & \textcolor{gray!75}{85.78} & \textcolor{gray!75}{43.31} & \textcolor{gray!75}{89.10} & \textcolor{gray!75}{76.99} & \textcolor{gray!75}{94.65} & \textcolor{gray!75}{98.74} 
        \\
        & SOR \cite{zhou2019dupnet} & \textbf{91.33} & 2.55 & 79.46 & 38.33 & 76.30 & 69.73 & \textbf{24.39} & 72.65 & 51.78 & 80.88 & 99.96 
        \\
        \cmidrule{2-13}
        & AT \cite{Madry18adversarial} & 86.91 & 3.08 & 65.96 & 42.79 & 61.75 & 72.93 & 26.34 & 71.80 & 60.62 & 74.11 & 98.14 
        \\
        & TRADES \cite{zhang2019trades} & 90.40 & 2.96 & 65.24 & 41.45 & 57.17 & 69.94 & 30.27 & 67.42 & 65.40 & 73.62 & 97.08 
        \\
        & LBGAT \cite{cui2020lbgat} & 84.00 & 3.08 & 40.60 & 27.39 & 49.26 & \textbf{48.82} & 33.75 & 64.67 & 66.17 & 71.64 & 97.02 
        \\
        & \textbf{Ours} & 91.25 & \colorbox{blue!20}{\textbf{2.27}} & \colorbox{blue!20}{\textbf{38.98}} & \colorbox{blue!20}{\textbf{21.31}} & \colorbox{blue!20}{\textbf{44.37}} & 69.25 & 25.41 & \colorbox{blue!20}{\textbf{50.00}} & \colorbox{blue!20}{\textbf{47.37}} & \colorbox{blue!20}{\textbf{51.34}} & \colorbox{blue!20}{\textbf{96.96}} 
        \\
        \midrule
        \midrule
        \multirow{6}{*}{CurveNet \cite{xiang2021curvenet}}
        & \textcolor{gray!75}{None} & \textcolor{gray!75}{93.80} & \textcolor{gray!75}{3.63} & \textcolor{gray!75}{92.46} & \textcolor{gray!75}{71.72} & \textcolor{gray!75}{92.22} & \textcolor{gray!75}{97.53} & \textcolor{gray!75}{42.87} & \textcolor{gray!75}{84.28} & \textcolor{gray!75}{79.17} & \textcolor{gray!75}{89.75} & \textcolor{gray!75}{99.92} 
        \\
        & SOR \cite{zhou2019dupnet} & \textbf{90.96} & 3.04 & 72.97 & 41.21 & 83.71 & 92.63 & \textbf{22.81} & 65.36 & 51.42 & 71.92 & 99.59 
        \\
        \cmidrule{2-13}
        & AT \cite{Madry18adversarial} & 89.26 & 3.04 & 46.56 & 31.24 & 51.99 & 67.46 & 30.96 & 55.43 & 50.16 & 60.45 & 92.06 
        \\
        & TRADES \cite{zhang2019trades} & 90.84 & 3.04 & 61.10 & 41.65 & 58.43 & 74.80 & 42.10 & 64.55 & 70.71 & 74.11 & 94.04 
        \\
        & LBGAT \cite{cui2020lbgat} & 85.74 & 2.84 & 28.77 & 20.18 & 42.33 & \textbf{40.72} & 35.58 & 56.36 & 60.53 & 61.18 & 92.94
        \\
        & \textbf{Ours} & 90.84 & \colorbox{blue!20}{\textbf{2.76}} & \colorbox{blue!20}{\textbf{24.96}} & \colorbox{blue!20}{\textbf{13.13}} & \colorbox{blue!20}{\textbf{36.67}} & 83.59 & 24.19 & \colorbox{blue!20}{\textbf{43.03}} & \colorbox{blue!20}{\textbf{42.54}} & \colorbox{blue!20}{\textbf{44.29}} & \colorbox{blue!20}{\textbf{90.11}} 
        \\
        \bottomrule
    \end{tabular}
    }
\end{table*}

\subsubsection{Datasets}

We conduct experiments on four datatsets (\ie, three popular point cloud object benchmarks and our proposed LiMN20) to comprehensively validate the performance of point cloud recognition. 
\begin{itemize}
    \item ModelNet40 \cite{Wu2015modelnet} consists of 12,311 CAD models from 40 man-made object categories, split into 9,843 for training and 2,468 for testing. Each point cloud is formed by 1,024 points which are uniformly sampled from the surface of each object and rescaled into a unit cube. 
    \item ShapeNetPart \cite{Yi16shapenet} contains 16,881 pre-aligned shapes from 16 categories that are more closer to the real LiDAR data, split into 12,137 for training and 2,874 for testing. We follow the same operations as ModelNet40 to process each point cloud sample.
    \item ModelNet40-C \cite{sun2022benchmarking} is a dataset specially designed for the corruption robustness of point cloud recognition. It is a corrupted version of ModelNet40 validation set that covers 15 common corruption types (\eg, occlusion, shearing or background noises) with 5 severity levels for each type.
    \item LiMN20 is our newly proposed dataset for verifying the recognition robustness under LiDAR scanning scenario. To simulate the LiDAR noise, we use a virtual Velodync HDL-64E2 scanner provided by Blensor \cite{Gschwandtner11blensor} to scan 100 3D meshes randomly selected from 20 confusing categories of ModelNet40. It totally contains 1,000 shapes, split into a half for ``easy'' set and the other half for ``hard'' set. The ``easy'' split is sampled by the standard simulated LiDAR scanner, while the ``hard'' split is sampled by the noisy simulated LiDAR scanner. More details can be found in \sref{limn20}.
\end{itemize}

\subsubsection{Models}
We generally consider four categories of point cloud recognition models to evaluate the proposed method, including PointNet \cite{charles2017pointnet}, PointNet++ \cite{charles2017pointnet++}, DGCNN \cite{wang2019dgcnn} and CurveNet \cite{xiang2021curvenet}. These models are really representative since they leverage very different strategies to learn the point features and greatly inspire the community.

\subsubsection{Implementation Details} \label{imp_detail}
Here we provide the default settings and hyper-parameters of PointCAT implementation. 
The learning rate is assigned as 0.001 and we use Adam optimizer with a cosine annealing schedule to train 155 epochs for all models. 
Besides, we fix a set of temperature hyper-parameters in all PointCAT training, \ie, $\tau_{adv}$, $\tau_{sup}$ and $\tau_{cen}$ are 0.1, 0.1 and 0.25 respectively.
For PointNet, the weight parameter $\alpha$ and $\beta$ are set as 8 and 0.5, respectively. 
For other recognition models, we unify $\alpha$ and $\beta$ as 1 and 4, respectively. 
Aiming for efficient training, the prototype update iteration number $T_1$ is assigned as 10 for inference-fast models (\ie, PointNet and DGCNN) and 2 for inference-slow models (\ie, PointNet++ and CurveNet). 
Moreover, the inner loop number of updating the noise generator is unified as 4 for all PointCAT training.
We ablate the aforementioned hyper-parameters in \sref{hyper}.

\begin{table*}[t]
	\footnotesize
    \centering
    \caption{Quantitative comparison about regular white-box robustness on ShapeNetPart, tested on PointNet (a.k.a, PN), PointNet++ (a.k.a, PN++), DGCNN and CurveNet. ``None'' denotes the recognition model without any defense. ``Acc'' means the accuracy on clean point clouds, higher is better. ``ASR'' means the attack success rate, lower is better.}
    \label{tab:table6}
    \setlength{\tabcolsep}{1mm}{
    \begin{tabular}{c|l|c|p{11mm}<{\centering}p{11mm}<{\centering}p{11mm}<{\centering}p{11mm}<{\centering}p{11mm}<{\centering}|p{11mm}<{\centering}p{11mm}<{\centering}p{11mm}<{\centering}p{11mm}<{\centering}p{11mm}<{\centering}}
        \toprule
        \multicolumn{1}{c|}{\multirow{3}{*}{Model}}
        & \multicolumn{1}{c|}{\multirow{3}{*}{Defense}}
        & \multicolumn{1}{c|}{\multirow{3}{*}{Acc (\%)}} 
        & \multicolumn{5}{c|}{ASR (Targeted Attack) (\%)} 
        & \multicolumn{5}{c}{ASR (Untargeted Attack) (\%)}
        \\
        & &
        & FGM
        & IFGM 
        & MIFGM 
        & PGD 
        & C\&W 
        & FGM 
        & IFGM 
        & MIFGM 
        & PGD 
        & C\&W 
        \\
        & &
        & \cite{goodfellow2015explaining}
        & \cite{gu2015towards}
        & \cite{dong18mifgm}
        & \cite{Madry18adversarial}
        & \cite{carlini2017towards}
        & \cite{goodfellow2015explaining}
        & \cite{gu2015towards}
        & \cite{dong18mifgm}
        & \cite{Madry18adversarial}
        & \cite{carlini2017towards}
        \\
        \midrule
        \multirow{6}{*}{PN \cite{charles2017pointnet}}
        & \textcolor{gray!75}{None} & \textcolor{gray!75}{98.71} & \textcolor{gray!75}{9.85} & \textcolor{gray!75}{94.71} & \textcolor{gray!75}{94.43} & \textcolor{gray!75}{95.37} & \textcolor{gray!75}{97.22} & \textcolor{gray!75}{53.62} & \textcolor{gray!75}{87.40} & \textcolor{gray!75}{93.49} & \textcolor{gray!75}{89.18} & \textcolor{gray!75}{86.15}
        \\
        & SOR \cite{zhou2019dupnet} & 98.58 & 4.40 & 33.88 & 17.19 & 51.14 & 85.09 & \textbf{5.18} & 18.11 & \textbf{16.48} & 19.67 & 60.23
        \\
        \cmidrule{2-13}
        & AT \cite{Madry18adversarial} & 98.36 & 6.72 & 62.25 & 60.02 & 66.21 & 79.61 & 32.71 & 60.06 & 64.89 & 62.56 & 45.34
        \\
        & TRADES \cite{zhang2019trades} & 98.43 & 7.17 & 72.06 & 69.62 & 76.97 & 82.67 & 28.53 & 66.98 & 71.19 & 69.52 & 52.99
        \\
        & LBGAT \cite{cui2020lbgat} & 96.80 & 6.02 & 35.14 & 33.99 & 36.99 & 45.20 & 28.22 & 65.69 & 71.99 & 69.07 & 41.75 
        \\
        & \textbf{Ours} & \colorbox{blue!20}{\textbf{98.61}} & \colorbox{blue!20}{\textbf{2.61}} & \colorbox{blue!20}{\textbf{7.97}} & \colorbox{blue!20}{\textbf{8.49}} & \colorbox{blue!20}{\textbf{10.26}} & \colorbox{blue!20}{\textbf{18.02}} & 6.78 & \colorbox{blue!20}{\textbf{17.71}} & 17.78 & \colorbox{blue!20}{\textbf{18.41}} & \colorbox{blue!20}{\textbf{24.84}} 
        \\
        \midrule
        \midrule
        \multirow{6}{*}{PN++ \cite{charles2017pointnet++}}
        & \textcolor{gray!75}{None} & \textcolor{gray!75}{99.03} & \textcolor{gray!75}{5.22} & \textcolor{gray!75}{85.00} & \textcolor{gray!75}{78.57} & \textcolor{gray!75}{86.36} & \textcolor{gray!75}{40.54} & \textcolor{gray!75}{17.54} & \textcolor{gray!75}{57.55} & \textcolor{gray!75}{55.43} & \textcolor{gray!75}{65.94} & \textcolor{gray!75}{33.89} 
        \\
        & SOR \cite{zhou2019dupnet} & 98.89 & 3.69 & 35.56 & 18.58 & 50.45 & 39.91 & 3.62 & 17.95 & 13.22 & 21.16 & 40.50
        \\
        \cmidrule{2-13}
        & AT \cite{Madry18adversarial} & 95.09 & 1.70 & 15.03 & \textbf{8.35} & 14.58 & 21.40 & 12.21 & 56.89 & 22.34 & 56.82 & 70.46 
        \\
        & TRADES \cite{zhang2019trades} & 98.33 & 4.49 & 25.71 & 18.23 & 37.89 & 35.94 & 5.88 & 18.44 & 27.28 & 28.25 & 20.88 
        \\
        & LBGAT \cite{cui2020lbgat} & 95.23 & 3.93 & 14.27 & 12.32 & 19.38 & 19.40 & 15.97 & 26.83 & 29.68 & 30.62 & \textbf{15.41}
        \\
        & \textbf{Ours} & \colorbox{blue!20}{\textbf{99.20}} & \colorbox{blue!20}{\textbf{1.43}} & \colorbox{blue!20}{\textbf{11.31}} & 8.73 & \colorbox{blue!20}{\textbf{13.99}} & \colorbox{blue!20}{\textbf{18.72}} & \colorbox{blue!20}{\textbf{2.51}} & \colorbox{blue!20}{\textbf{11.27}} & \colorbox{blue!20}{\textbf{10.09}} & \colorbox{blue!20}{\textbf{12.70}} & 23.56 
        \\
        \midrule
        \midrule
        \multirow{6}{*}{DGCNN \cite{wang2019dgcnn}}
        & \textcolor{gray!75}{None} & \textcolor{gray!75}{99.10} & \textcolor{gray!75}{7.55} & \textcolor{gray!75}{81.21} & \textcolor{gray!75}{60.13} & \textcolor{gray!75}{83.61} & \textcolor{gray!75}{86.88} & \textcolor{gray!75}{14.16} & \textcolor{gray!75}{58.94} & \textcolor{gray!75}{61.38} & \textcolor{gray!75}{67.71} & \textcolor{gray!75}{97.08}  
        \\
        & SOR \cite{zhou2019dupnet} & 98.71 & 4.83 & 32.29 & 12.21 & 47.15 & 69.14 & 8.63 & 29.19 & 16.91 & 33.16 & 87.82 
        \\
        \cmidrule{2-13}
        & AT \cite{Madry18adversarial} & 98.19 & 5.36 & 28.25 & 19.73 & 35.14 & 66.25 & 10.68 & 42.80 & 47.70 & 48.33 & 88.34 
        \\
        & TRADES \cite{zhang2019trades} & 98.57 & 6.05 & 32.43 & 19.00 & 45.02 & 77.52 & 8.18 & 31.46 & 34.94 & 35.70 & 70.94
        \\
        & LBGAT \cite{cui2020lbgat} & 96.49 & 4.59 & 30.02 & 18.09 & 34.69 & 61.27 & 11.69 & 29.96 & 38.17 & 36.64 & 38.90
        \\
        & \textbf{Ours} & \colorbox{blue!20}{\textbf{98.99}} & \colorbox{blue!20}{\textbf{2.92}} & \colorbox{blue!20}{\textbf{7.83}} & \colorbox{blue!20}{\textbf{5.11}} & \colorbox{blue!20}{\textbf{12.63}} & \colorbox{blue!20}{\textbf{30.72}} & \colorbox{blue!20}{\textbf{5.01}} & \colorbox{blue!20}{\textbf{10.09}} & \colorbox{blue!20}{\textbf{11.13}} & \colorbox{blue!20}{\textbf{11.93}} & \colorbox{blue!20}{\textbf{29.92}} 
        \\
        \midrule
        \midrule
        \multirow{6}{*}{CurveNet \cite{xiang2021curvenet}}
        & \textcolor{gray!75}{None} & \textcolor{gray!75}{98.94} & \textcolor{gray!75}{9.32} & \textcolor{gray!75}{80.69} & \textcolor{gray!75}{61.00} & \textcolor{gray!75}{86.36} & \textcolor{gray!75}{97.18} & \textcolor{gray!75}{34.20} & \textcolor{gray!75}{57.52} & \textcolor{gray!75}{63.26} & \textcolor{gray!75}{62.18} & \textcolor{gray!75}{80.41} 
        \\
        & SOR \cite{zhou2019dupnet} & 98.33 & 4.03 & 15.42 & 5.21 & 26.67 & 76.53 & 12.92 & 17.57 & 15.97 & 17.78 & 50.07 
        \\
        \cmidrule{2-13}
        & AT \cite{Madry18adversarial} & 98.85 & 6.26 & 32.74 & 21.89 & 42.97 & 65.27 & 12.67 & 24.60 & 29.71 & 28.60 & 24.22 
        \\
        & TRADES \cite{zhang2019trades} & 98.68 & 7.06 & 40.15 & 27.49 & 52.85 & 78.50 & 16.53 & 37.06 & 43.39 & 41.16 & 31.42 
        \\
        & LBGAT \cite{cui2020lbgat} & 98.40 & 5.46 & 13.08 & 10.51 & 17.12 & \textbf{29.89} & 20.22 & 35.91 & 38.73 & 38.73 & \textbf{20.91}
        \\
        & \textbf{Ours} & \colorbox{blue!20}{\textbf{98.99}} & \colorbox{blue!20}{\textbf{2.92}} & \colorbox{blue!20}{\textbf{7.83}} & \colorbox{blue!20}{\textbf{5.11}} & \colorbox{blue!20}{\textbf{12.63}} & 30.72 & \colorbox{blue!20}{\textbf{5.01}} & \colorbox{blue!20}{\textbf{10.09}} & \colorbox{blue!20}{\textbf{11.13}} & \colorbox{blue!20}{\textbf{11.93}} & 29.92 
        \\
        \bottomrule
    \end{tabular}
    }
\end{table*}

\subsection{Robustness on Adversarial Attacks}

\subsubsection{Regular White-box Attacks}
We compare our method with the following baselines, \ie, point cloud defense SOR \cite{zhou2019dupnet} and the most advanced adversarial training methods including PGD-based AT \cite{Madry18adversarial}, TRADES ($1/\lambda=1$) \cite{zhang2019trades} and LBGAT ($\alpha=0$) \cite{cui2020lbgat}. 
Note that all adversarial training baselines share the same training settings with our PointCAT, including the same learning rate as 0.001, the same perturbation threshold as 0.04 and the same inner loop number as 4. 
Also we adopt the default hyper-parameter setting of SOR. 
To verify the white-box robustness of point cloud models equipped with these defenses, we conduct regular gradient-based adversarial attacks FGM \cite{goodfellow2015explaining}, IFGM \cite{gu2015towards}, MIFGM \cite{dong18mifgm}, PGD \cite{Madry18adversarial} and optimization-based attack C\&W \cite{carlini2017towards} from both targeted and untargeted perspectives. 

For targeted gradient-based attacks, the attack iterations are assigned as 50 and the perturbation threshold is 0.08 under the $l_2$-norm constraint. 
For targeted C\&W attack, we set 10 binary steps with total 500 iterations for the adversarial point cloud optimization, in which the learning rate is 0.01. 
For untargeted gradient-based attacks, the attack iterations are 10 and the threshold is 0.02 also with under $l_2$-norm constraint. 
For untargeted C\&W attack, we adopt 5 binary steps with total 250 iterations, where the optimizing learning rate is 0.003. 
The step size of all these attacks is calculated by dividing $\delta\sqrt{K\times C}$ by the iteration number, where $\delta$ is the perturbation threshold and $K\times C$ denotes the input dimensions of each point cloud. 

\begin{table}[t]
	\footnotesize
    \centering
    \caption{Robust accuracy (\%) under targeted attacks on PointNet. For clearer comparison, we use the weaker attack setting than Table \ref{tab:table1} and \ref{tab:table6}. Our method still achieves the best.}
    \label{tab:table_ra}
    \setlength{\tabcolsep}{4mm}{
    \begin{tabular}{l|cccc}
    \toprule
    \multirow{2}{*}{Defense}
    & IFGM 
    & MIFGM 
    & PGD 
    & C\&W 
    \\
    & \cite{gu2015towards}
    & \cite{dong18mifgm}
    & \cite{Madry18adversarial}
    & \cite{carlini2017towards}
    \\
    \midrule
    \textcolor{gray!75}{None} & \textcolor{gray!75}{1.50} & \textcolor{gray!75}{2.31} & \textcolor{gray!75}{1.18} & \textcolor{gray!75}{0.69} 
    \\
    SOR \cite{zhou2019dupnet} & 32.46 & 31.89 & 27.67 & 8.55 
    \\
    AT \cite{Madry18adversarial} & 15.60 & 15.28 & 10.82 & 22.89 
    \\
    TRADES \cite{zhang2019trades} & 14.59 & 11.87 & 10.74 & 13.45 
    \\
    LBGAT \cite{cui2020lbgat} & 18.07 & 19.33 & 16.65 & 28.61 
    \\
    \textbf{Ours} & \textbf{43.23} & \textbf{38.17} & \textbf{35.66} & \textbf{30.23} 
    \\
    \bottomrule
    \end{tabular}
    }
\end{table}

\begin{figure*}[!h]
\centering
\begin{minipage}{0.245\linewidth}
    \centering
    \includegraphics[width=1\linewidth]{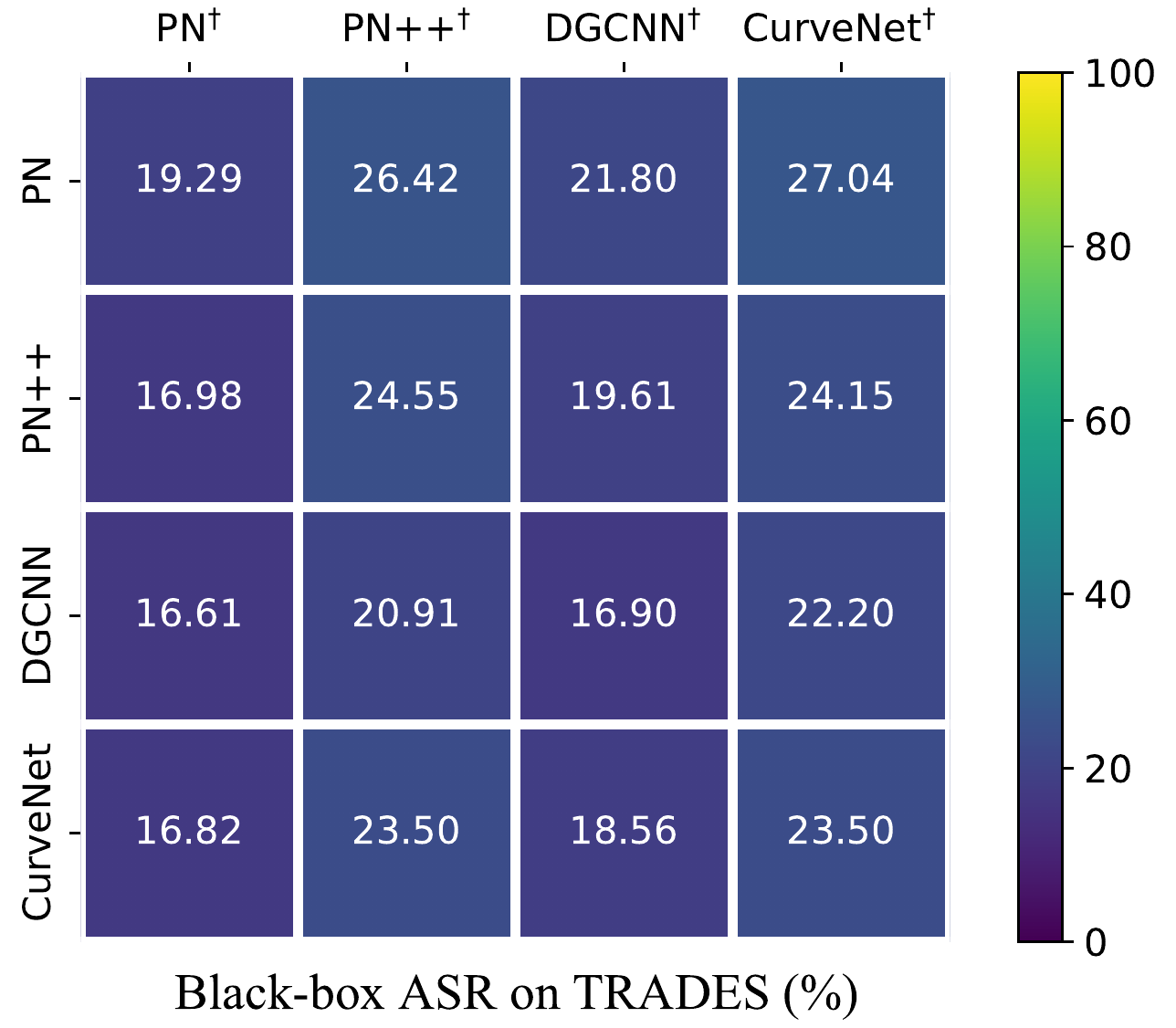}
\end{minipage}
\hfill
\begin{minipage}{0.245\linewidth}
    \centering
    \includegraphics[width=1\linewidth]{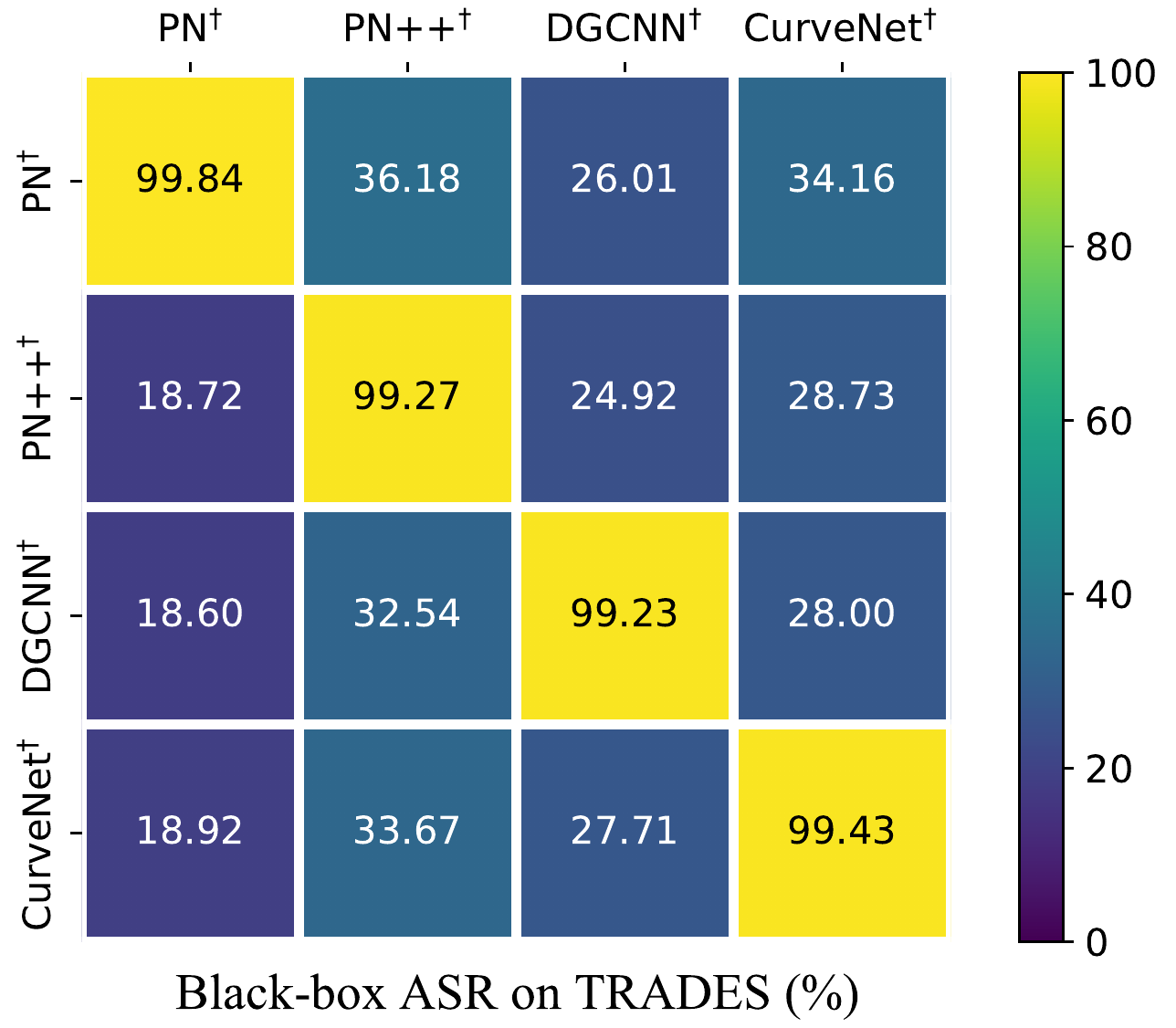}
\end{minipage}
\hfill
\begin{minipage}{0.245\linewidth}
    \centering
    \includegraphics[width=1\linewidth]{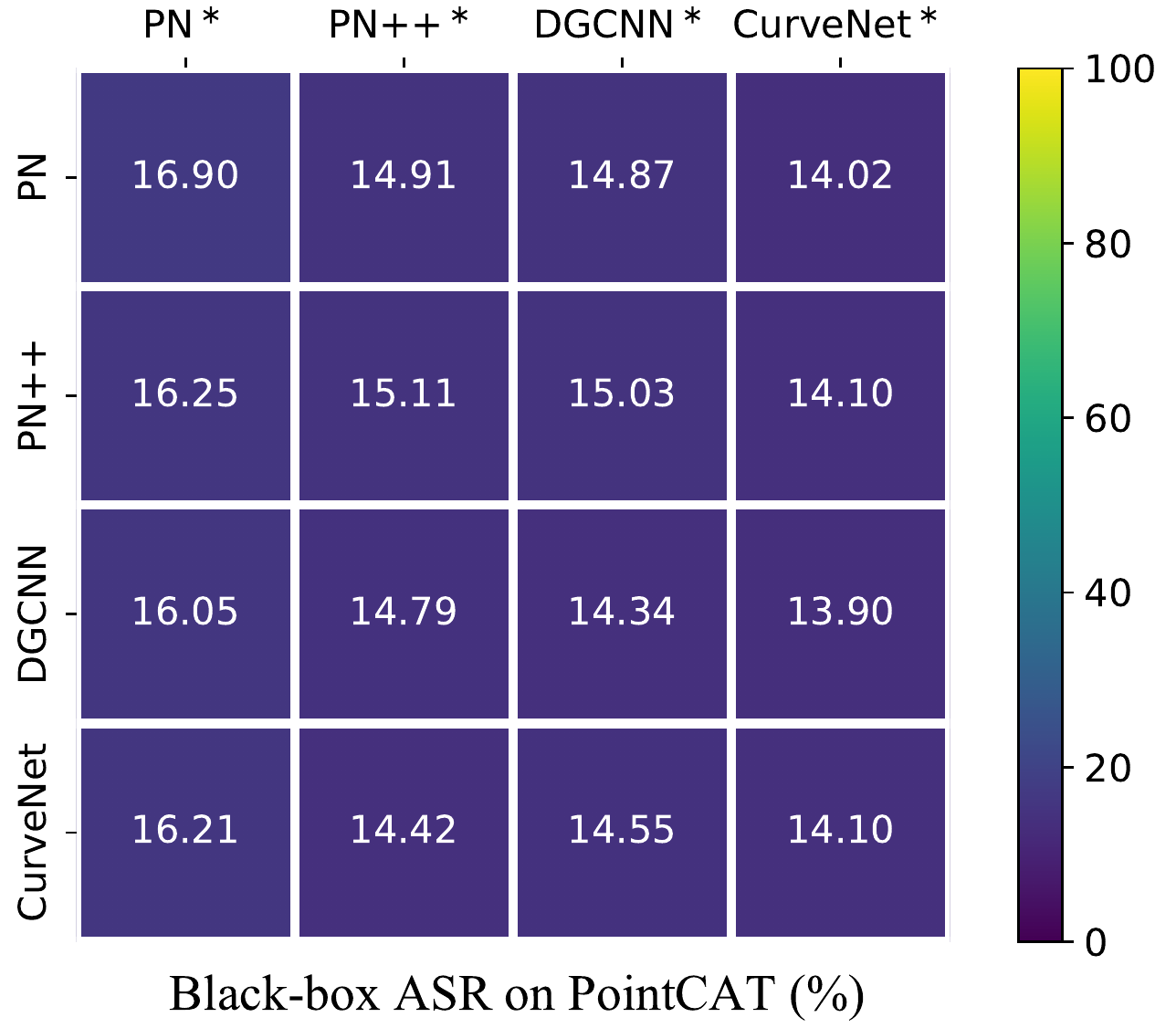}
\end{minipage}
\hfill
\begin{minipage}{0.245\linewidth}
    \centering
    \includegraphics[width=1\linewidth]{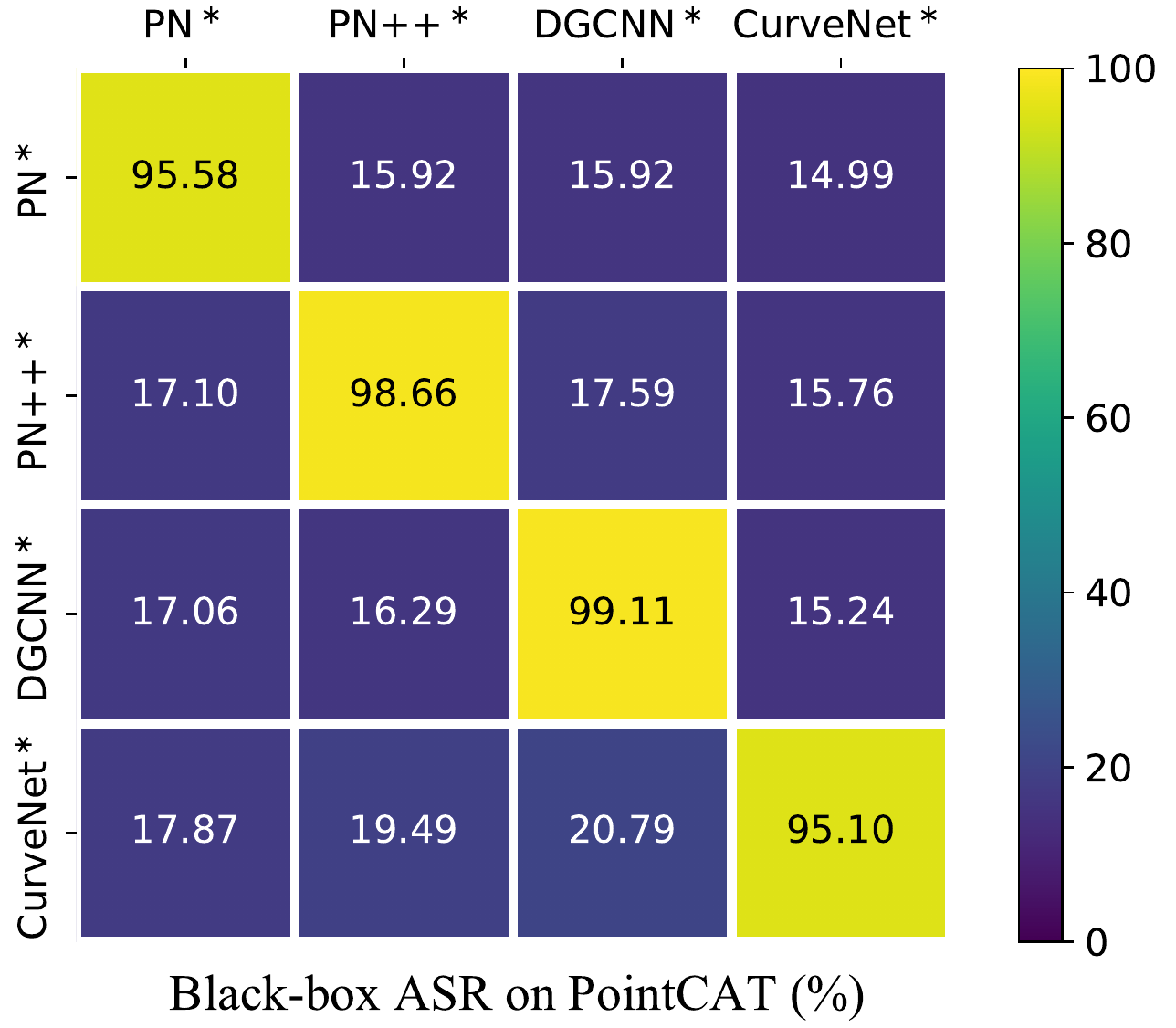}
\end{minipage}
\vfill
\vspace{1em}
\begin{minipage}{0.245\linewidth}
    \centering
    \includegraphics[width=1\linewidth]{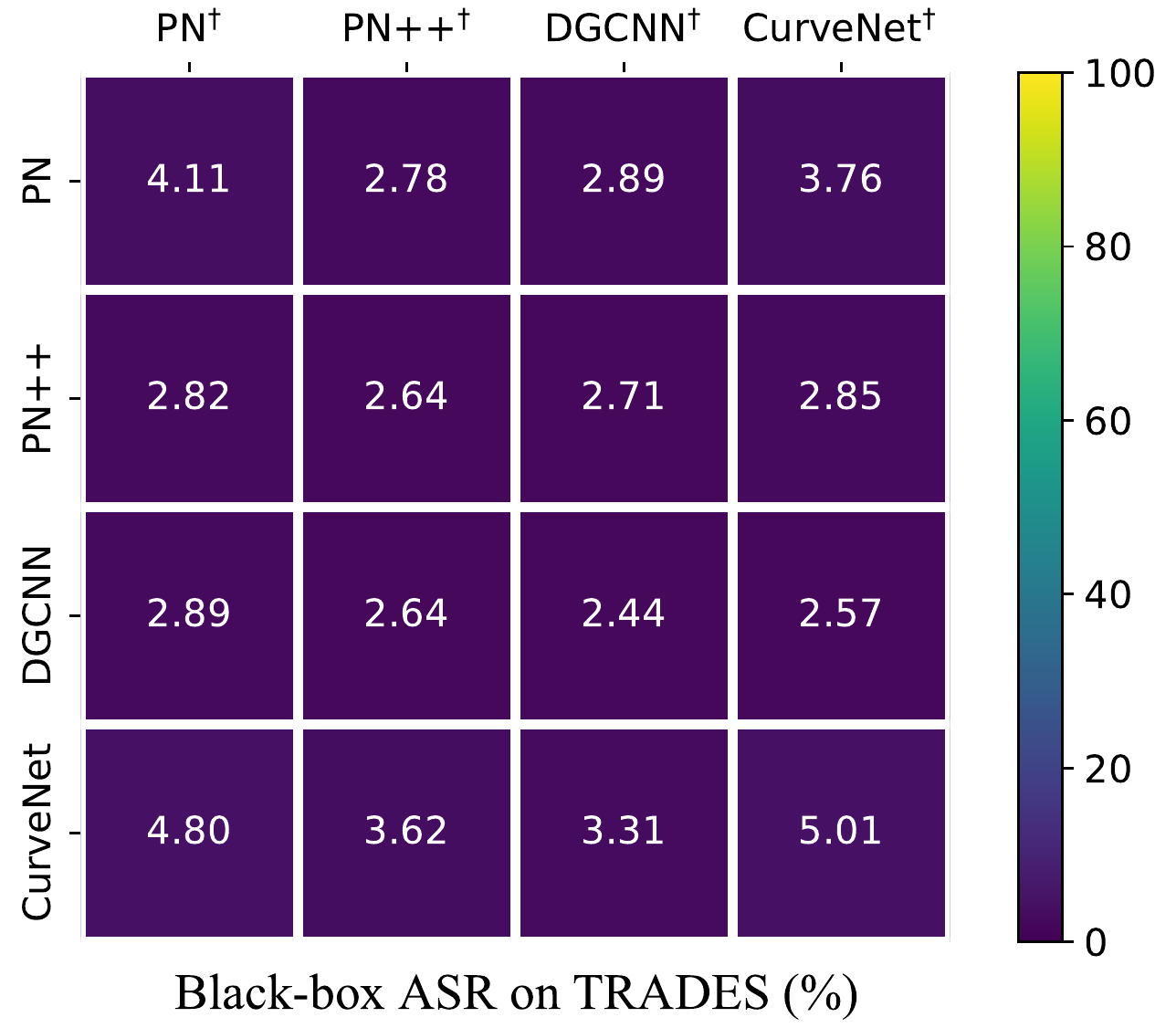}
\end{minipage}
\hfill
\begin{minipage}{0.245\linewidth}
    \centering
    \includegraphics[width=1\linewidth]{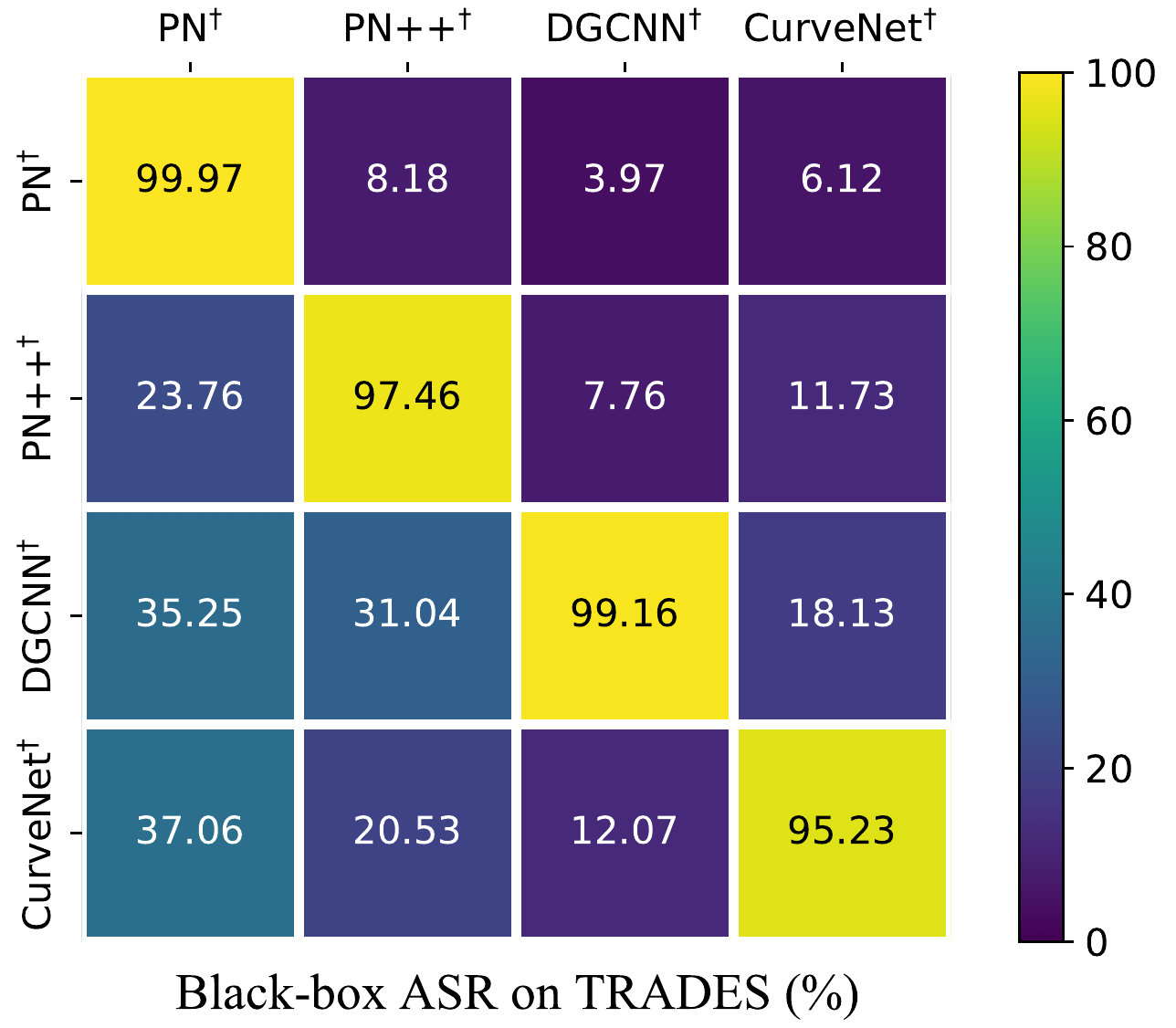}
\end{minipage}
\hfill
\begin{minipage}{0.245\linewidth}
    \centering
    \includegraphics[width=1\linewidth]{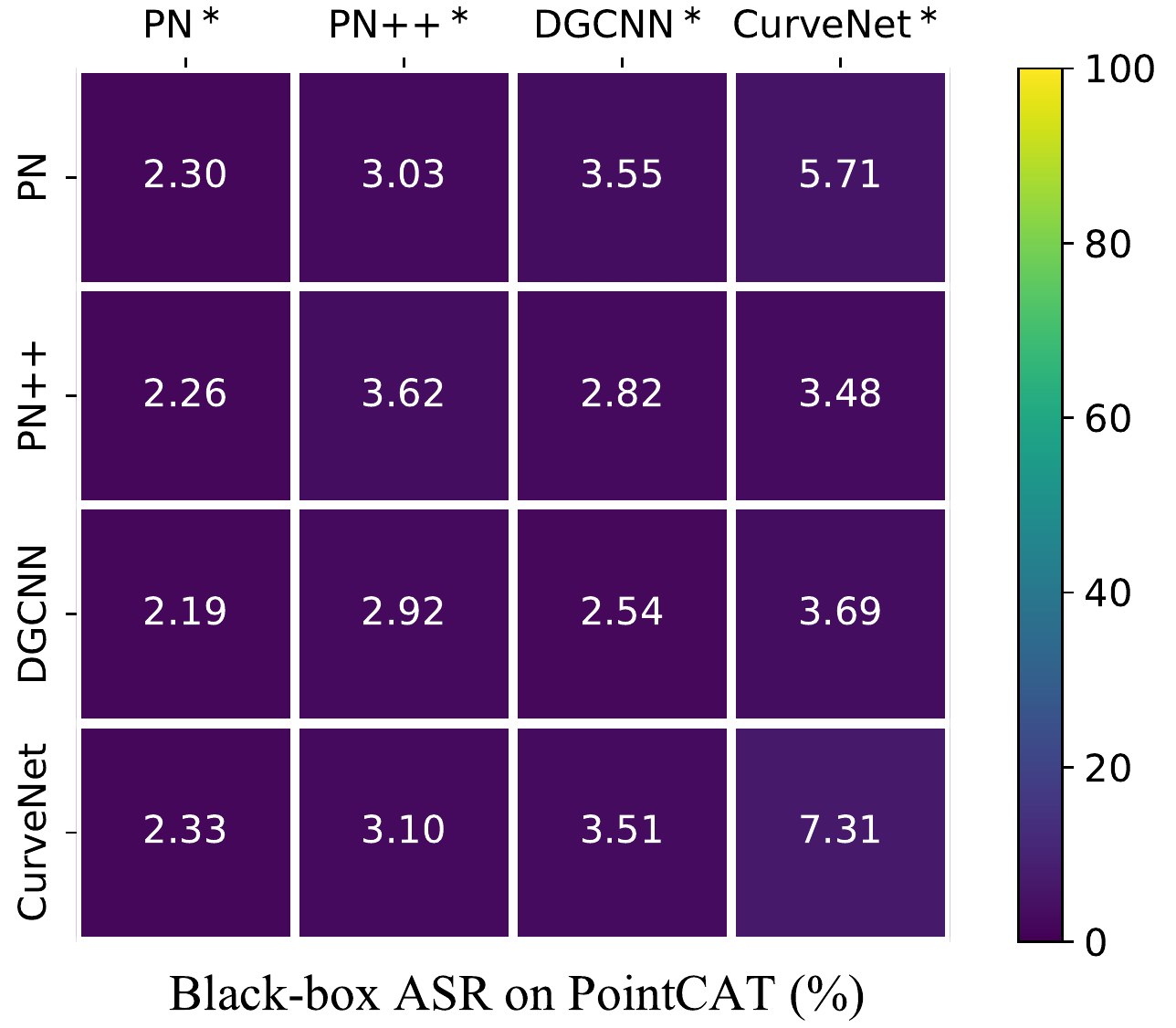}
\end{minipage}
\hfill
\begin{minipage}{0.245\linewidth}
    \centering
    \includegraphics[width=1\linewidth]{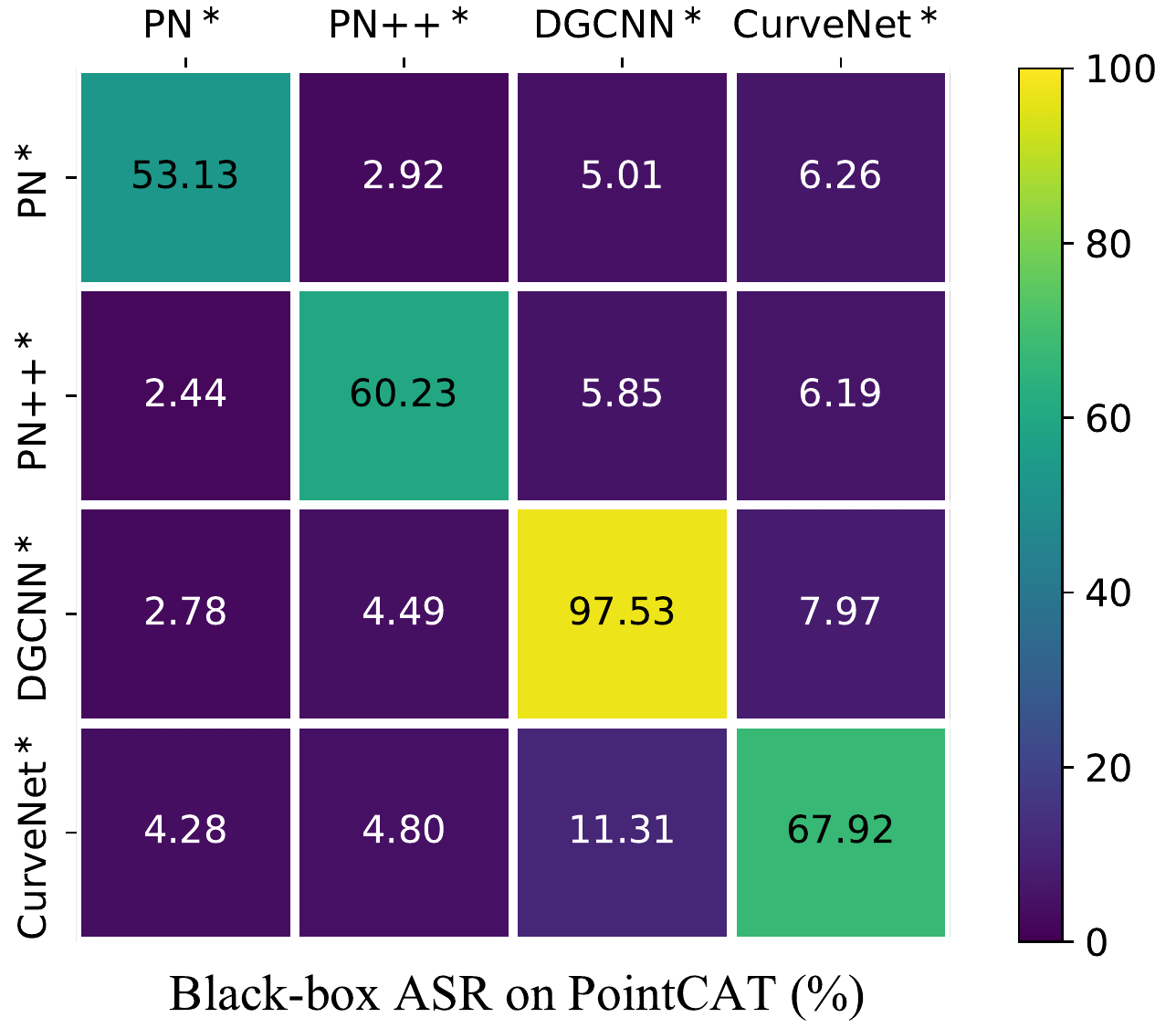}
\end{minipage}
\caption{Black-box tranferability attack success rate (ASR) results for TRADES and PointCAT on ModelNet40 (\textbf{the top four figures}) and ShapeNetPart (\textbf{the bottom four figures}). In each figure, values in the same row correspond to the same source model used as the white-box victim, while values in the same column correspond to the same target model. Here $\dag$ means the model trained by TRADES and $\ast$ means the model trained by PointCAT.}
\label{fig:heatmap}
\end{figure*}

As the comprehensive results shown in Table \ref{tab:table1} and \ref{tab:table6}, the point cloud recognition models trained by our PointCAT performs generally more robust when confronted with different regular white-box adversarial attacks, with really few clean accuracy degraded. 
It can be easily noticed that other adversarial training methods which succeed in image recognition tasks are still limited for point cloud recognition: 
TRADES maintains comparable clean accuracy but gets weaker adversarial robustness than PointCAT, while LBGAT obtains better robustness than TRADES but sacrifices more on clean accuracy. 
Another surprising finding is that, in Table \ref{tab:table6} (ShapeNetPart dataset results), the clean accuracy (Acc) of PointNet++ and CurveNet trained by our method is even higher than the vanilla clean baseline (denoted as ``None''). 
Considering that, in targeted attack scenario, simply calculating ASR to measure whether the model is robust is not entirely convincing, we also report the model classification accuracy (namely robust accuracy) in Table \ref{tab:table_ra} and derive the similar conclusion. 
Overall, our PointCAT achieves the best performance on both clean accuracy and model robustness.

\begin{table}[t]
	\footnotesize
    \centering
    \caption{Quantitative comparison on attack success rate (ASR) of Auto-Attack and advanced point cloud adversarial attacks. The listed defenses are implemented on PointNet.}
    \label{tab:table2}
    \setlength{\tabcolsep}{1mm}{
    \begin{tabular}{l|cccc}
    \toprule
    Defense
    & AA \cite{croce2020autoattack}
    & 3D-Adv \cite{Xiang3dadv}
    & AdvPC \cite{HamdiRTG20advpc}
    & GeoA$^3$ \cite{geoa3}
    \\
    \midrule
    \textcolor{gray!75}{None} & \textcolor{gray!75}{95.30} & \textcolor{gray!75}{99.88} & \textcolor{gray!75}{99.84} & \textcolor{gray!75}{100.00}
    \\
    AT \cite{Madry18adversarial} & 65.19 & 94.81 & 78.61 & 99.60
    \\
    TRADES \cite{zhang2019trades} & 66.69 & 98.14 & 88.94 & 99.96
    \\
    \textbf{Ours} & \textbf{63.90} & \textbf{37.07} & \textbf{57.54} & \textbf{74.27}
    \\
    \bottomrule
    \end{tabular}
    }
\end{table}

\begin{table}[t]
	\footnotesize
    \centering
    \caption{Quantitative comparison on ModelNet40 with different settings implemented on CurveNet (a.k.a, CN), for the robust accuracy on isotropic Gaussian noisy or sparse point clouds. ``$T_{in}$'' denotes the average inference time.}
    \label{tab:table3}
    \setlength{\tabcolsep}{1mm}{
    \begin{tabular}{l|c|cc|cc|c}
    \toprule
    \multicolumn{1}{l|}{\multirow{2}{*}{Robust Setting}}
    & \multicolumn{1}{c|}{\multirow{2}{*}{Acc (\%)}}
    & \multicolumn{2}{c|}{Noisy Acc (\%)}
    & \multicolumn{2}{c|}{Sparse Acc (\%)}
    & \multicolumn{1}{c}{\multirow{2}{*}{$T_{in}$ (s)}}
    \\
    \cmidrule{3-4}
    \cmidrule{5-6}
    &  & 4\% & 8\% & 70\% & 80\% &
    \\
    \midrule
    CN \cite{xiang2021curvenet} & 93.80 & 68.88 & 10.66 & 73.58 & 49.59 & 0.17
    \\
    \midrule
    \multicolumn{7}{l}{\textit{Equiped with outlier removal}}
    \\
    CN+SOR \cite{zhou2019dupnet} & \textbf{90.96} & 76.46 & 27.84 & 61.35 & 38.86 & 0.21
    \\
    CN+DUP-Net \cite{zhou2019dupnet} & 87.88 & 75.16 & 37.03 & 65.52 & 43.07 & 1.32
    \\
    \midrule
    \multicolumn{7}{l}{\textit{Equiped with denoising or upsampling}}
    \\
    CN+DMR \cite{luo2020dmr} & 75.89 & 77.67 & 63.86 & - & - & 0.65
    \\
    CN+PU-Net \cite{yu2018punet} & 88.25 & 67.54 & 16.49 & 84.44 & 68.60 & 1.32
    \\
    \midrule
    \multicolumn{7}{l}{\textit{Equiped with adversarial training}}
    \\
    CN (AT \cite{Madry18adversarial}) & 89.26 & 81.20 & 47.49 & 88.90 & 87.32 & 0.17
    \\
    CN (TRADES \cite{zhang2019trades}) & 90.84 & 77.76 & 38.74 & 88.86 & 87.48 & 0.17
    \\
    \textbf{CN (Ours)} & 90.84 & \textbf{88.01} & \textbf{63.98} & \textbf{89.95} & \textbf{88.25} & \textbf{0.17}
    \\
    \bottomrule
    \end{tabular}
    }
\end{table}

\subsubsection{White-box Auto-Attack (AA) and Point Cloud Attacks}
To further evaluate the efficiency of PointCAT on more stronger white-box attacks, we implement Auto-Attack \cite{croce2020autoattack} and three recently proposed point cloud adversarial attacks, \ie, 3D-Adv \cite{Xiang3dadv}, AdvPC \cite{HamdiRTG20advpc} and GeoA$^3$ \cite{geoa3} with the default settings elaborated in their papers. 
Auto-Attack is widely recognized as a reliable approach for model robustness evaluation, which ensembles APGD-CE, APGD-DLR, FAB and Square Attack. 
Only the first three are adopted here since Square Attack is specially designed for images and hard to be extended to point clouds. 
As the results listed in Table \ref{tab:table2}, the PointNet trained by PointCAT achieves the state-of-the-art robustness under these strong white-box attacks. 
Especially on three point cloud adversarial attacks, we dramatically outperforms both standard adversarial training and TRADES with lower ASR values. 

\subsubsection{Black-box Attacks}
We further verify the adversarial robustness of the proposed method on black-box attack. 
For the fairness of this comparison, all of these recognition models are trained on ModelNet40 (or ShapeNetPart) training split under the same settings with PointCAT. 
Specifically, we apply untargeted $FGSM^{50}$ (black-box) for both evaluation on two datasets, which has 50 iterations for gradient ascent and gets $l_2$-norm constrained. 
The step size setup is same with that adopted for white-box attacks. 
The adversarial point clouds are generated on the source model and tested on the target model as its input. 
As shown in \fref{fig:heatmap}, the black-box attack transferability is much lower among PointCAT trained models than TRADES trained models. 
Accordingly, PointCAT enables point cloud models to share the stronger black-box robustness with each other and better restrict with tranfer-based attacks.

\begin{figure*}[t]
\centering
\includegraphics[width=0.9\linewidth]{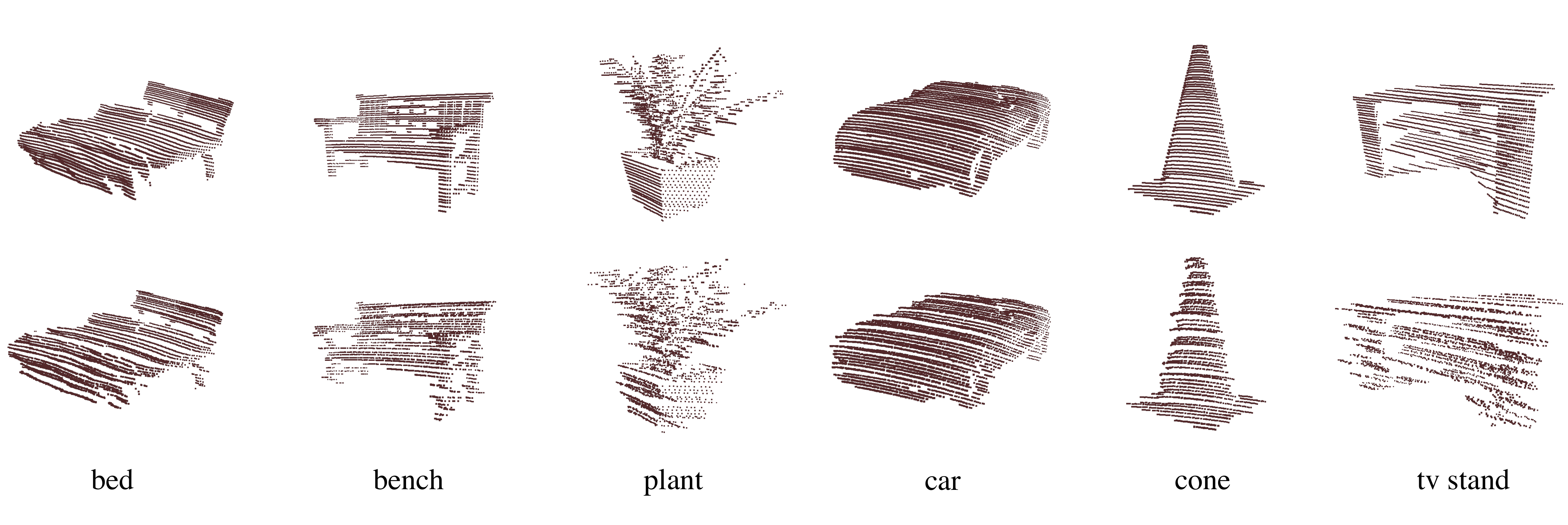}
\caption{Visualization for some samples selected from ``easy'' split (\textbf{top row}) and ``hard'' split (\textbf{bottom row}) of LiMN20.}
\label{fig:supp_3}
\end{figure*}

\begin{table}[t]
	\footnotesize
    \centering
    \caption{Overall recognition error rates on \textbf{ModelNet40-C}. The model we selected here is PointNet.}
    \label{tab:table_modelnetc}
    \setlength{\tabcolsep}{2.2mm}{
    \begin{tabular}{lccccc}
    \toprule
    & None
    & AT
    & TRADES
    & LBGAT
    & \cellcolor{blue!20}\textbf{Ours}
    \\
    \midrule
    Error rate (\%) & 26.3 & 25.3 & 26.0 & 32.0 & \cellcolor{blue!20}\textbf{25.2}
    \\
    \bottomrule
    \end{tabular}
    }
\end{table}

\begin{figure}[t]
\centering
\includegraphics[width=0.95\linewidth]{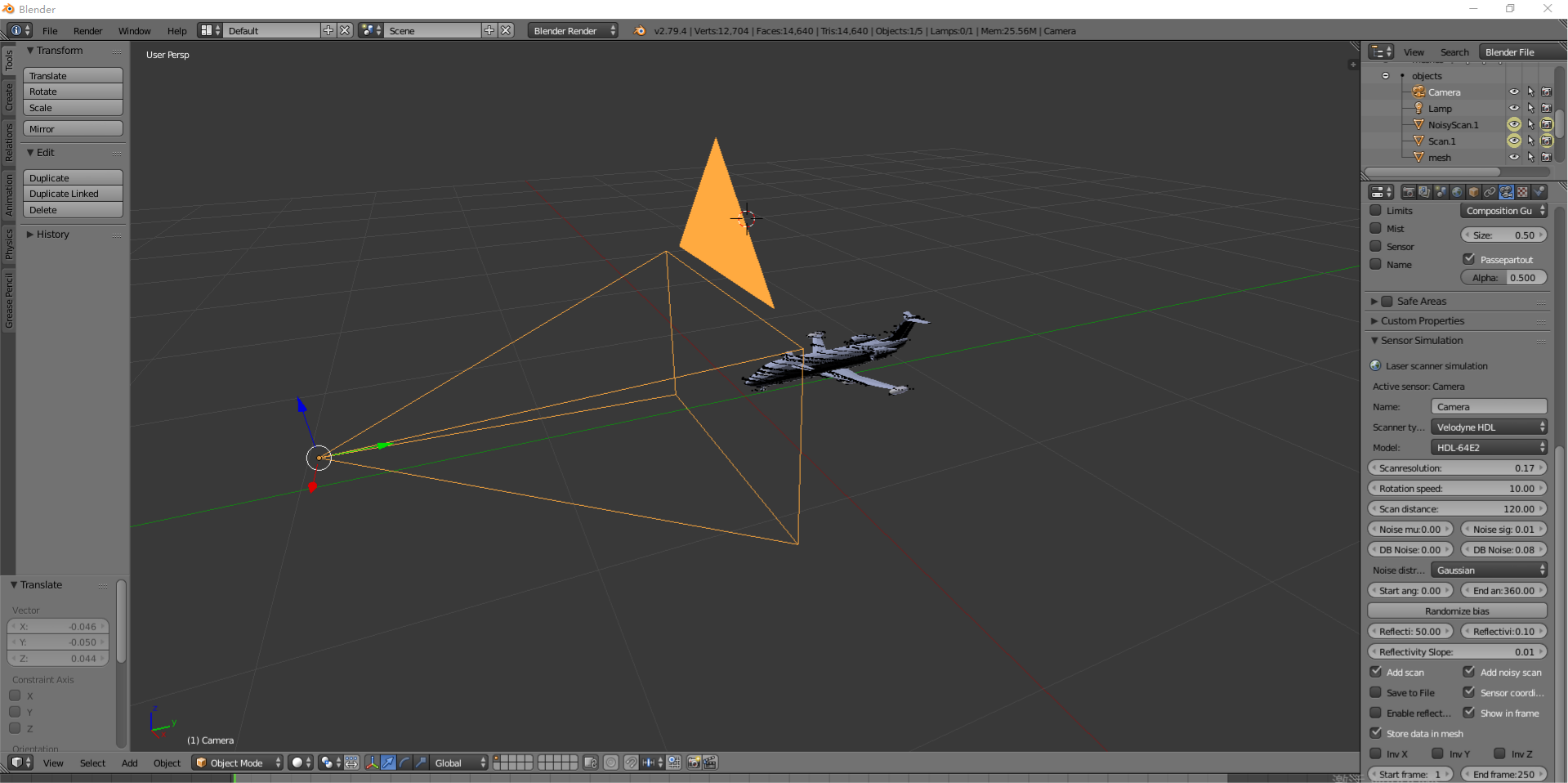}
\caption{The LiDAR scanning simulation with Blensor package.}
\label{fig:blensor}
\end{figure}

\subsection{Robustness on Natural Corruptions}

\subsubsection{Random Point Noise and Point Dropping}
Since the point clouds collected in the real-world are often mutilated or perturbed due to the complicated environments, it is necessary for the recognition model to resist the natural corruptions. 
In Table \ref{tab:table3}, we first utilize isotropic Gaussian noise and random point dropping to corrupt the model input of baselines and our models. 
The standard variance of Gaussian noise is assigned as 4\% or 8\% to mimic the point deviation from the surface, and the ratio of point dropping is set as 70\% or 80\% to construct sparse point clouds. 
We test different robust settings on CurveNet (a.k.a, CN), including 
1) its vanilla version; 
2) CN equipped with extra outlier removal modules SOR \cite{zhou2019dupnet} or DUP-Net \cite{zhou2019dupnet}; 
3) CN equipped with extra point cloud denoising module DMR \cite{luo2020dmr}; 
4) CN equipped with extra point cloud upsampling module PU-Net \cite{yu2018punet}; 
5) CN trained by adversarial training methods standard AT \cite{Madry18adversarial}, TRADES \cite{zhang2019trades} or our PointCAT. 
The results summarized in Table \ref{tab:table3} shows that our method can greatly outperform the baselines at both the robust accuracy and the inference efficiency.

\subsubsection{ModelNet40-C Common Corruptions}
ModelNet40-C \cite{sun2022benchmarking} is a recently proposed dataset for benchmarking the robustness of point cloud recognition models when confronting with different kinds of distortions. 
These distortions are common in real-world scenarios that are relevant with LiDAR scanning, which are divided into three categories, \ie, density, noise and transformation. 
In this paper, we directly run the recognition models on this benchmark to comprehensively verify their natural robustness. 
To make a fair comparison, we take the default configuration of ModelNet40-C during evaluation. 
The results are shown in Table \ref{tab:table_modelnetc}.
We can find that PointCAT obtains the lowest overall error rate compared with other adversarial training baselines, indicating the best performance.

\begin{figure}[t]
\centering
\includegraphics[width=0.8\linewidth]{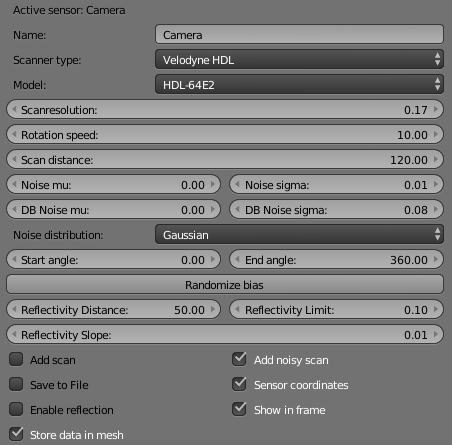}
\caption{Detailed settings used for noisy simulated LiDAR scanner in Blensor package.}
\label{fig:supp_4}
\end{figure}

\subsubsection{LiDAR Simulated Noise on LiMN20.} \label{limn20}
In the more realistic scenario, we should take the distortions caused by LiDAR perceptron into considerations and test the recognition models under such distortions. 
Previous 3D object datasets uniformly sample points on shapes and these points are noise-free, which are mismatched with rotary scanning used by LiDAR in real complicated scenarios. 
Therefore, we intend to contribute a new dataset named LiMN20 to fill this gap, which prepares for LiDAR-scanned point clouds. 
To simulate the LiDAR noise, we use a virtual Velodync HDL-64E2 scanner provided by Blensor \cite{Gschwandtner11blensor} to scan 100 3D meshes, which are randomly selected from 20 confusing categories (\eg, bookshelf and tv stand) of ModelNet40 \cite{Wu2015modelnet}. 
Sampled from different angles and positions with 1) the standard simulated LiDAR scanner and 2) the noisy simulated LiDAR scanner respectively, the proposed LiMN20 contains total 1,000 point clouds, in which 500 shapes construct the ``easy'' split and the other 500 shapes form the ``hard'' split. 
Moreover, the point number of each point cloud is varied from 1,000 to 8,000, simulating the uncertain quantity of reflected points in real-world scanning scenarios. 
The visualization of some samples, the virtual Blensor simulation \cite{Gschwandtner11blensor} of LiDAR scanners and the settings for configuring the noisy simulated LiDAR are provided in \fref{fig:supp_3}, \ref{fig:blensor} and \ref{fig:supp_4}, respectively.

\begin{table}[t]
	\footnotesize
    \centering
    \caption{Robust accuracy on the proposed dataset LiMN20. We verify CurveNet on both ``easy'' and ``hard'' validation splits. ``None'' means the vanilla model without any defenses.}
    \label{tab:table4}
    \setlength{\tabcolsep}{1mm}{
    \begin{tabular}{l|ccccc}
    \toprule
    & None
    & SOR \cite{zhou2019dupnet}
    & AT \cite{Madry18adversarial}
    & TRADES \cite{zhang2019trades}
    & \textbf{Ours}
    \\
    \midrule
    Acc (easy) & 48.60 & 45.40 & 42.00 & 37.20 & \textbf{65.00}
    \\
    Acc (hard) & 32.80 & 35.20 & 34.20 & 31.40 & \textbf{52.20}
    \\
    \bottomrule
    \end{tabular}
    }
\end{table}

To better mimic the corruptions brought by real-world LiDAR scanning, we further verify baselines and our model on the proposed dataset LiMN20. 
The evaluation results are listed in Table \ref{tab:table4}. 
Apparently, our method can also dramatically boost the accuracy under the simulated LiDAR noisy scenario, which demonstrates the practicality and the resiliency against the scanning noise.

\begin{figure}[t]
\centering
\begin{minipage}{0.49\linewidth}
    \centering
    \includegraphics[width=1\linewidth]{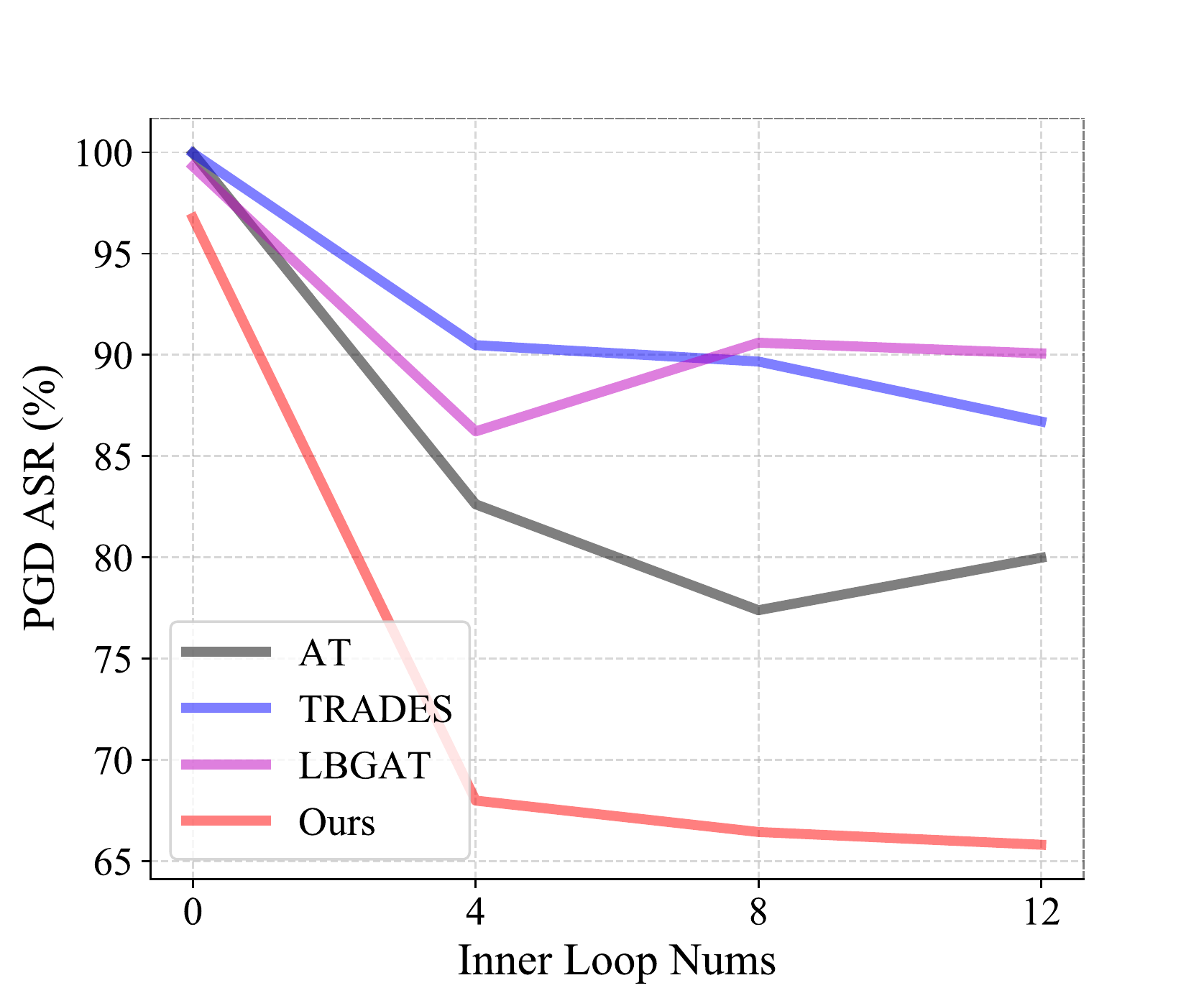}
\end{minipage}
\hfill
\begin{minipage}{0.49\linewidth}
    \centering
    \includegraphics[width=1\linewidth]{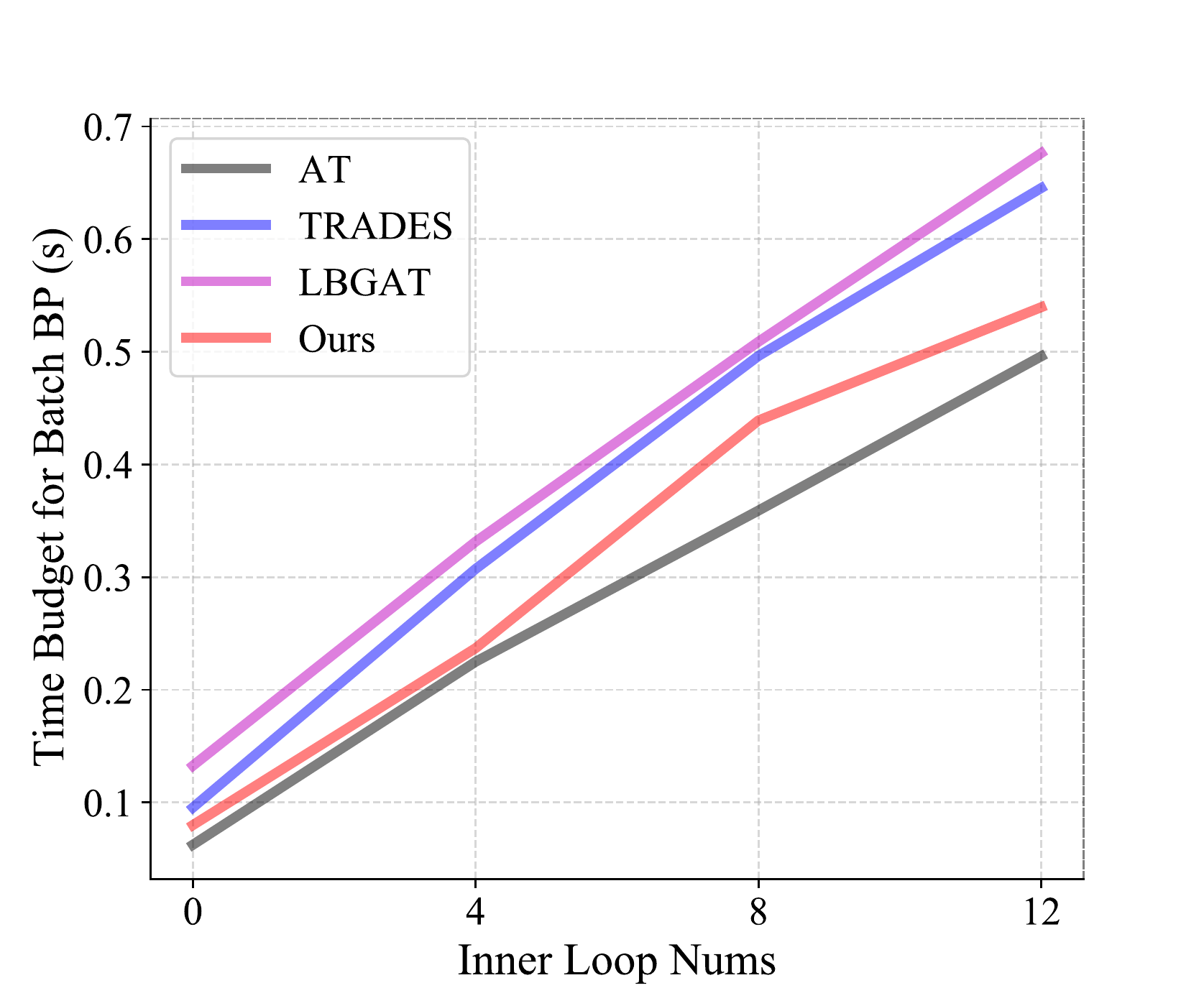}
\end{minipage}
\caption{Ablation studies for different inner loop numbers. We report attack success rate (ASR) of white-box PGD and the average time budget for each batch back-propagation (batch size is 16). The perturbation threshold is unified as 0.04.}
\label{fig:inner_loop_nums}
\end{figure}

\begin{figure}[t]
\vspace{-1.7em}
\centering
\includegraphics[width=1.0\linewidth]{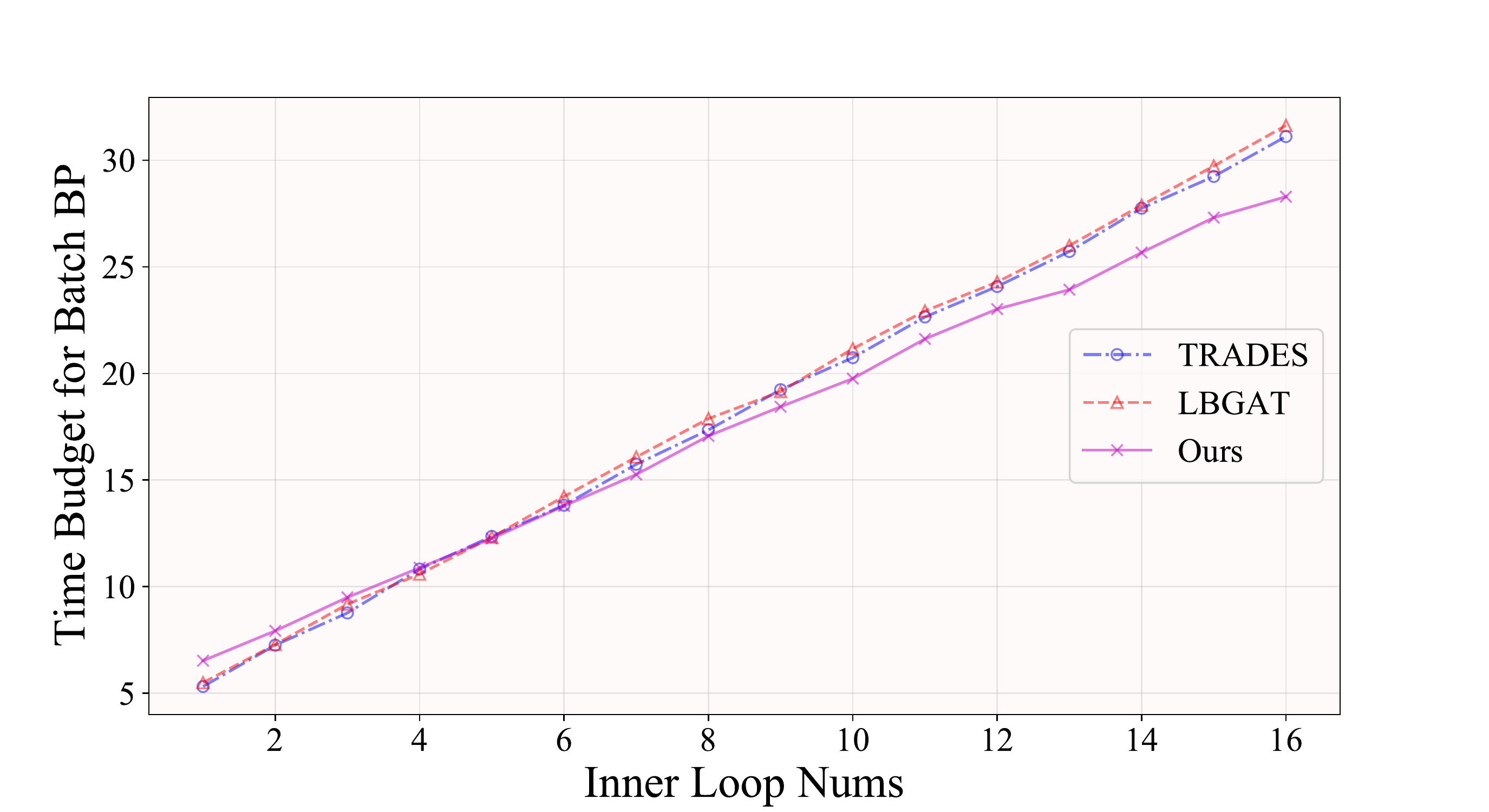}
\caption{Average time budget for batch data back-propagation. We test PointNet++ on different inner loops (batch size is 16).}
\label{fig:supp_1}
\end{figure}

\subsection{Ablation Studies}

\subsubsection{Different Inner Loop Numbers $T_2$} \label{loop}
To explore the robustness improvement brought by inner loop numbers, we conduct the evaluation on PointNet trained by baselines and our method for 0, 4, 8, 12 inner loops, respectively. 
Two aspects of comparison are considered here, \ie, the robustness against white-box PGD attack and the average time budget for back-propagation in each batch training (batch size is 16). 
As indicated in \fref{fig:inner_loop_nums}, our method outperforms the baseline methods a lot while inheriting the lower time budget than TRADES and LBGAT. 
Even implemented with only 4 inner loops, our method are also better than previous adversarial training using PGD-8 and PGD-12 (\ie, AT with 8, 12 inner loops). 
It proves that despite replacing the traditional PGD-based inner loop as adversarial noise generation, PointCAT is also able to boost more robustness for point cloud recognition than previous adversarial training methods that take the extra time cost on more inner loops.

\subsubsection{Running Time Budget Analysis} \label{time}

To clarify the time efficiency of PointCAT, we further conduct the experiments to obtain the average time cost for batch data back-propagation with TRADES \cite{zhang2019trades}, LBGAT \cite{cui2020lbgat} and PointCAT. 
For the fairness of evaluation, we adopt the same training settings, including the same perturbation threshold as 0.04, the same learning rate as 0.001, the same batch size as 16 and the same RTX 3090Ti GPU devices. 
Note that the trained model is unified as PointNet++ \cite{charles2017pointnet++}, and our method adopt the same prototype update settings with that is given in \sref{imp_detail}, \ie, $T_1 = 2$, $\eta_1 = 0.005$. 
The detailed results are showcased in \fref{fig:supp_1}. 
When configuring fewer inner loops, PointCAT is little more time-consuming due to the involvement of extra prototype computation. 
When configuring more inner loops, PointCAT achieves the better time efficiency than both TRADES and LBGAT, owing to the replacement of traditional PGD loops with the lightweight noise generator training.

\begin{table}[t]
	\footnotesize
    \centering
    \caption{Ablation Studies for choosing model inputs instead of features as variables to update prototypes. We report the natural robust accuracy (RA) against 8\% isotropic noises, 80\% point dropping and attack success rate (ASR) against targeted/untargeted PGD attack.}
    \label{tab:table_opt}
    \setlength{\tabcolsep}{1mm}{
    \begin{tabular}{l|c|cc|cc}
    \toprule
    \multicolumn{1}{l|}{\multirow{2}{*}{Variable}}
    & \multicolumn{1}{c|}{\multirow{2}{*}{Acc (\%)}}
    & \multicolumn{2}{c|}{Natural RA (\%) $\uparrow$}
    & \multicolumn{2}{c}{PGD ASR (\%) $\downarrow$}
    \\
    \cmidrule{3-4}
    \cmidrule{5-6}
    &  & Noise(8\%) & Drop(80\%) & Targeted & Untargeted
    \\
    \midrule
    feature & 87.52 & 64.26 & 85.70 & 72.57 & 73.82
    \\
    model input & 87.97 & 67.54 & 86.18 & 24.07 & 67.99
    \\
    \bottomrule
    \end{tabular}
    }
\end{table}

\subsubsection{Different Ways for Prototype Update}

As formulated in Eq.(\ref{eq:3}), we optimize model inputs to realize the data-independent prototype update. 
While a more straightforward way is directly optimizing hypersphere features and just computing on the last classification layer. 
But unfortunately, the results in \Tref{tab:table_opt} shows that this way is much less effective than optimizing model inputs.
It is because that directly optimizing features is unrestricted while optimizing model inputs just allows features to be confined to a specific encoding distribution.

\subsubsection{Importance of The Proposed Mechanisms} \label{importance}
To clarify the significance of two loss components $\mathcal{L}_{sup}$, $\mathcal{L}_{cen}$, the adversarial noise generation and the dynamic prototype guidance, we conduct the ablation study on each of them. 
For two loss components, we abandon either of them in Eq.(\ref{eq:robust_loss}) to test the performance when just leveraging the remaining loss component. 
For adversarial noise generation, we replace it with the PGD-based inner loop to find the difference between before and after. 
For the prototype guidance, we remove the centralizing loss in Eq.(\ref{eq:robust_loss}) and the escaping component in Eq.(\ref{eq:adv_loss}). 
As indicated in Table \ref{tab:table5}, all of these losses or mechanisms can dramatically help our method boost the adversarial robustness of point cloud recognition models. 
The results also shows that, especially when equipping with $\mathcal{L}_{cen}$, the prototype guidance takes an essential part in overall robustness improvement. 
It is consistent with the intuition of avoiding the learned features of positive pairs deviating from the ground-truth category cluster, where the corrupted positive samples are learning to be more challenging.

\begin{figure}[t]
\centering
\vspace{-1em}
\begin{minipage}{0.492\linewidth}
    \centering
    \includegraphics[width=1\linewidth]{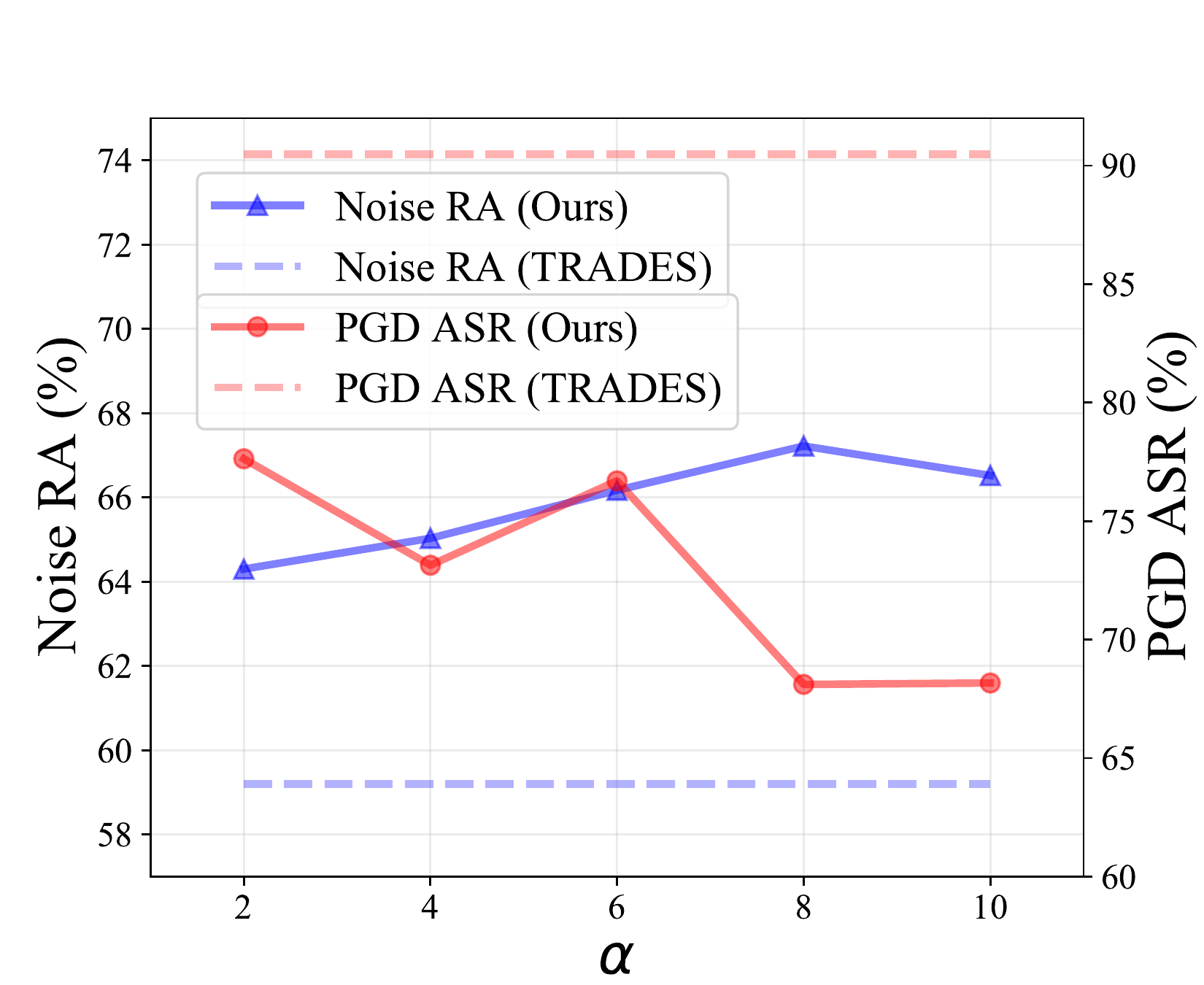}
\end{minipage}
\hfill
\begin{minipage}{0.492\linewidth}
    \centering
    \includegraphics[width=1\linewidth]{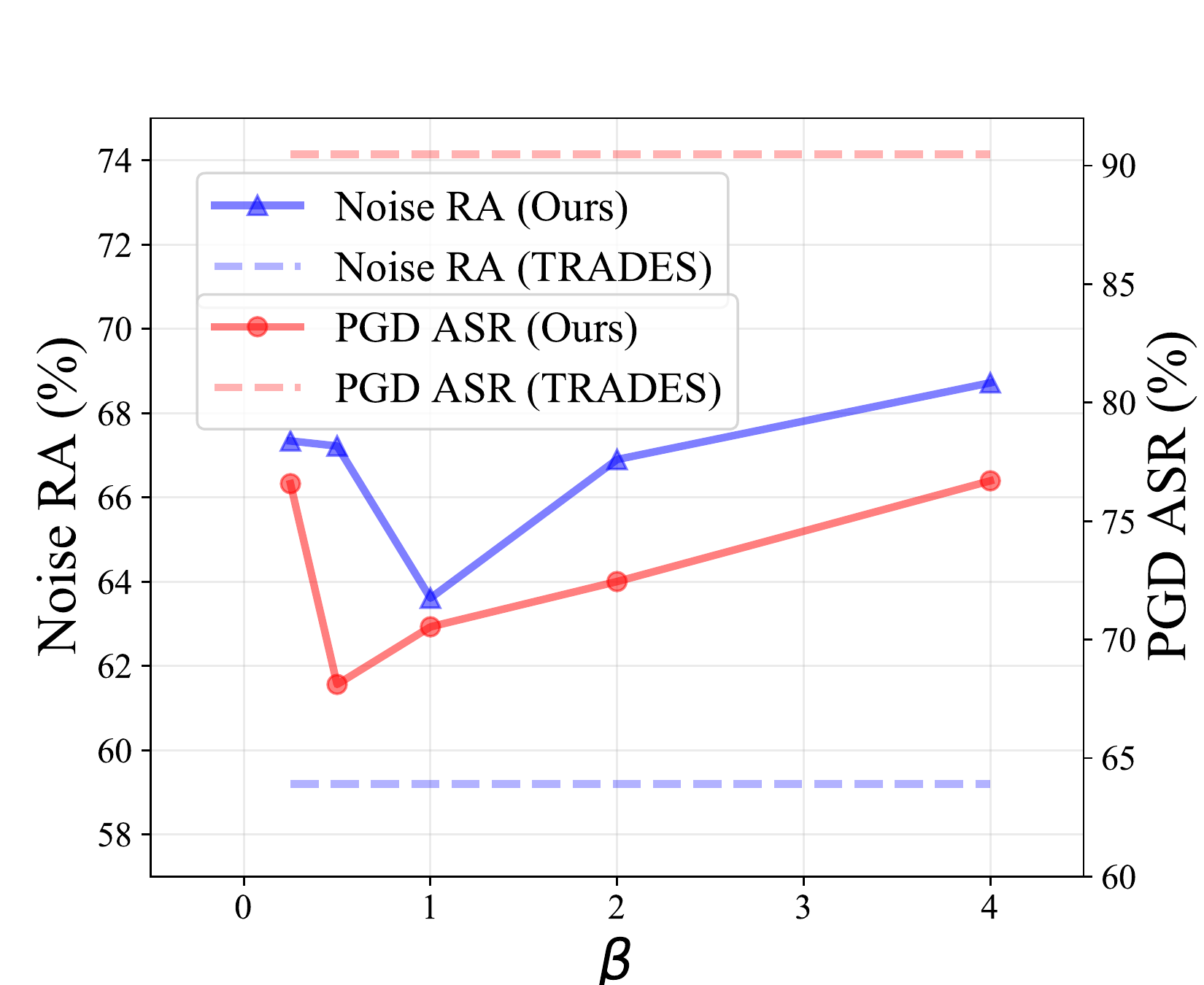}
\end{minipage}
\caption{Ablation studies for hyper-parameters $\alpha$ and $\beta$. We report the natural robust accuracy (RA) against 8\% isotropic noises and attack success rate (ASR) against untargeted PGD attack. We select $\alpha=8$, $\beta=0.5$ on PointNet eventually.}
\label{fig:figure_abl}
\end{figure}

\begin{table}[t]
	\footnotesize
    \centering
    \caption{Ablation Studies for two loss components $\mathcal{L}_{sup}$, $\mathcal{L}_{cen}$, noise generator $\mathcal{G}$ and the prototype guidance.}
    \label{tab:table5}
    \setlength{\tabcolsep}{1mm}{
    \begin{tabular}{l|c|cc|cc}
    \toprule
    \multicolumn{1}{l|}{\multirow{2}{*}{Setting}}
    & \multicolumn{1}{c|}{\multirow{2}{*}{Acc (\%)}}
    & \multicolumn{2}{c|}{Natural RA (\%) $\uparrow$}
    & \multicolumn{2}{c}{PGD ASR (\%) $\downarrow$}
    \\
    \cmidrule{3-4}
    \cmidrule{5-6}
    &  & Noise(8\%) & Drop(80\%) & Targeted & Untargeted
    \\
    \midrule
    w/o $\mathcal{L}_{sup}$ & 86.63 & 60.22 & 86.02 & 37.80 & 71.43 
    \\
    w/o $\mathcal{L}_{cen}$ & 87.88 & 61.13 & 85.90 & 60.01 & 90.15 
    \\
    w/o $\mathcal{G}$ & 87.86 & 62.12 & 85.67 & 53.97 & 78.00 
    \\
    w/o prototype & 87.82 & 64.51 & 86.06 & 61.99 & 91.05
    \\
    \midrule
    \textbf{Ours} & \textbf{87.97} & \textbf{67.54} & \textbf{86.18} & \textbf{24.07} & \textbf{67.99} 
    \\
    \bottomrule
    \end{tabular}
    }
\end{table}

\begin{table}[t]
	\footnotesize
    \centering
    \caption{Ablation Studies for three temperature hyper-parameters $\tau_{adv}$, $\tau_{sup}$ and $\tau_{cen}$ in our loss functions. The results explains that why we unify $\tau_{adv}=0.1$, $\tau_{sup}=0.1$, $\tau_{cen}=0.25$ in all of PointCAT training.}
    \label{tab:table_abl}
    \setlength{\tabcolsep}{1mm}{
    \begin{tabular}{l|c|cc|cc}
    \toprule
    \multicolumn{1}{l|}{\multirow{2}{*}{Setting}}
    & \multicolumn{1}{c|}{\multirow{2}{*}{Acc (\%)}}
    & \multicolumn{2}{c|}{Natural RA (\%) $\uparrow$}
    & \multicolumn{2}{c}{PGD ASR (\%) $\downarrow$}
    \\
    \cmidrule{3-4}
    \cmidrule{5-6}
    &  & Noise(8\%) & Drop(80\%) & Targeted & Untargeted
    \\
    \midrule
    $\tau_{adv}=0.07$ & 87.12 & 65.84 & 85.45 & 25.26 & 74.76 
    \\
    $\tau_{adv}=0.13$ & \textbf{88.21} & 67.22 & 85.82 & 24.23 & 73.46 
    \\
    \midrule
    $\tau_{sup}=0.07$ & 87.52 & 64.99 & 86.02 & 29.54 & 74.59  
    \\
    $\tau_{sup}=0.13$ & 87.84 & \textbf{68.64} & 85.82 & 24.55 & 71.64  
    \\
    \midrule
    $\tau_{cen}=0.20$ & 87.93 & 66.00 & 85.58 & 27.31 & 68.11 
    \\
    $\tau_{cen}=0.30$ & 87.28 & 67.46 & 85.78 & 25.61 & 74.88 
    \\
    \midrule
    \textbf{Ours} & \textbf{87.97} & \textbf{67.54} & \textbf{86.18} & \textbf{24.07} & \textbf{67.99} 
    \\
    \bottomrule
    \end{tabular}
    }
\end{table}

\subsubsection{Ablating Hyper-Parameters} \label{hyper}

The configuration of the introduced hyper-parameters has a significant impact on the model robustness. 
To investigate such impact, we conduct a series of ablation experiments on different hyper-parameter settings. 
As the results listed in \fref{fig:figure_abl} and Table \ref{tab:table_abl}. 
Though hyper-parameters change, our method always outperforms baselines with its relatively robust performance. 
With the unified implementation details given in \sref{imp_detail}, it is applicable for implement our method on most of common point cloud models to outperform existing baselines. 
Moreover, it definitely provides more flexible ways for achieving the more satisfying performance by fine tuning these hyper-parameters for the specific point cloud recognition model.

\subsubsection{Efficiency vs. Effectiveness}
The prototype guidance is an integral part of the performance improvement as discussed in \sref{importance}. 
While the dynamic prototype update mechanism requires more computation cost for training, we should clarify that such extra time budget is limited and configurable during training. 
First, we have shown that the overall running time of PointCAT is generally comparable with other adversarial training methods in \sref{loop} and \sref{time}. 
Second, we further verify that a modest reduction in the prototype update times (\ie, fewer update iterations $T_1$) does not degrade the robustness a lot. 
Specifically, we can reduce the prototype update times to boost the training efficiency with little performance sacrificed, \eg, when we replace $T_1 = 10$, $\eta_1 = 0.001$ (10 iterations, 0.001 update rate) with $T_1 = 2$ , $\eta_1 = 0.005$ (2 iterations , 0.005 update rate) for DGCNN \cite{wang2019dgcnn}, the degradation of its adversarial robustness is only 0.37\%  on untargeted PGD \cite{Madry18adversarial} attack. 
Therefore, ``more prototype update iterations'' is an optional enhancement for further improving the robustness if the computation resources are sufficient.

\section{Conclusion}

In this paper, we propose \textbf{P}oint-\textbf{C}loud \textbf{C}ontrastive \textbf{A}dversarial \textbf{T}raining \textbf{(PointCAT)}, to boost the general robustness of point cloud object recognition. 
To facilitate the category-wise alignment and the uniformity of learned features on the hypersphere, we specially design a pair of centralizing losses and supervised contrastive loss for recognition model training. 
With the purpose of online searching the more challenging corrupted point clouds, a noise generator is adversarially training along with the recognition model from the scratch. 
Extensive verification on ModelNet40, ShapeNetPart, ModelNet40-C and LiMN20 for various point cloud recognition models demonstrate that, our method achieves the superior performance against both white-box and black-box adversarial perturbations including strong Auto-Attack, and natural corruptions such as isotropic point noise, point dropping and the simulated LiDAR noise. 
Besides, a new dataset named LiMN20 is contributed in this paper, for validating the recognition robustness under the simulated LiDAR noisy scanning environment.

\ifCLASSOPTIONcaptionsoff
  \newpage
\fi

\bibliographystyle{IEEEtran}
\bibliography{IEEEabrv,egbib}

\end{document}